\documentclass[10pt,letterpaper,twoside]{article}

\usepackage[utf8]{inputenc}
\usepackage{fancyhdr}
\usepackage{amsmath}
\usepackage{amssymb}
\usepackage{booktabs}
\usepackage{makecell}
\usepackage{graphicx}
\usepackage{bm}
\usepackage{svg}
\usepackage{tabularx}
\usepackage{subcaption}
\usepackage{stfloats}
\usepackage[hidelinks]{hyperref}
\usepackage{multirow}
\usepackage{xcolor}
\usepackage{siunitx}
\usepackage{float}
\usepackage{listings}
\usepackage{setspace}
\usepackage{url}
\usepackage{xspace}
\usepackage{colortbl}
\usepackage[bb=stixtwo]{mathalpha}
\usepackage[nameinlink,capitalize,noabbrev]{cleveref}
\crefname{methods}{Methods}{Methods}
\Crefname{methods}{Methods}{Methods}
\crefname{appendix}{Appendix}{Appendices}
\Crefname{appendix}{Appendix}{Appendices}
\usepackage[font=small,labelfont=bf,labelsep=period]{caption}
\usepackage[letterpaper,portrait,bottom=0.7in,top=1in,left=0.7in,right=0.7in]{geometry}
\usepackage[square,numbers,sort&compress]{natbib}
\usepackage[scaled=0.95]{inter}

\usepackage[scaled=1.03]{inconsolata}
\usepackage{ulem}

\usepackage[T1]{fontenc}
\usepackage{helvet}

\newcommand{\xvr}{\textbf{\textit{xvr}}\xspace}

\hypersetup{colorlinks=true, linkcolor=blue, urlcolor=blue, citecolor=blue}
\fancypagestyle{firstpage}{%
  \fancyhf{}%
  \fancyfoot[L]{\footnotesize \textsuperscript{\ensuremath{\dagger}} Contact: vbutoi@mit.edu, ndey@mgh.harvard.edu}%
}

\newcommand{\radcite}[2]{%
  \begin{minipage}{\linewidth}
    \noindent\hangindent=2em\hangafter=1 #1 \\
    \noindent\textit{#2}
    \vspace{0.75em}
  \end{minipage}
}

\newcommand{\subpara}[1]{\vspace{0.4em}\noindent\textit{\uline{#1.}}\hspace{0mm}}
\newcommand{\paragraphsection}[1]{\paragraph{#1}\mbox{}\par}

\newcommand{\method}{FleXray\xspace}
\newcommand{\lightrule}{\arrayrulecolor{black!20}\midrule[0.3pt]\arrayrulecolor{black}}

\renewenvironment{abstract}
{
\list{}{
\setlength{\leftmargin}{0mm}
\setlength{\rightmargin}{0mm}
}
\item\relax
}
{\endlist}

\makeatletter
\renewcommand{\maketitle}{
\bgroup\setlength{\parindent}{0pt}
\begin{flushleft}
{\fontsize{25pt}{25pt}\selectfont\textbf{\@title}}
\\\vspace{20pt}
\@author
\\
\@date
\end{flushleft}\egroup
}
\makeatother

\title{\method{}: Universal Clinical X-ray Segmentation}

\author{
\textbf{
Victor Ion Butoi~\textsuperscript{\ensuremath{\dagger},1},~
Vivek Gopalakrishnan~\textsuperscript{1},~
John V. Guttag~\textsuperscript{1},~
Adrian V.~Dalca~\textsuperscript{1,2,3},~
Neel Dey~\textsuperscript{\ensuremath{\dagger},1,2,3} \\
}
\vspace{0.5em}
{\small
\textbf{1.}~MIT CSAIL
\textbf{2.}~Massachusetts General Hospital
\textbf{3.}~Harvard Medical School
}}

\date{}

\begin{document}

\twocolumn[
\begin{@twocolumnfalse}
\maketitle

\thispagestyle{firstpage}

\begin{abstract}
\small
\textbf{Abstract.}
X-ray is medicine's most widely used imaging modality, yet remains among its least quantitative. Unlike volumetric modalities like CT or MRI, X-ray collapses 3D anatomy into a 2D projection, causing structures to overlap and anatomical boundaries to be ambiguous, even to experts. As a result, labeling X-ray databases for training general-purpose segmentation systems is impractical, leaving morphometric and functional X-ray analysis confined to narrow anatomical regions and applications. To this end, we present \method{}, a generalist model for anatomical segmentation across the entire body in clinical X-rays. Instead of curating large, manually annotated X-ray datasets, we build a scalable, physics-based generative X-ray data engine. Using existing 3D whole-body CT segmentation datasets and generative image-editing models, we simulate \textit{fully-annotated} 2D X-rays with diverse appearances, physiological properties, and imaging geometries. Trained on these simulations, \method{} accurately segments 60 anatomical structures across unseen research datasets and in-the-wild X-rays. We further show that \method{} makes X-rays directly amenable to quantitative analysis, enabling automated measurements for disease grading, robust navigation during X-ray-guided interventions, and data-efficient learning of pathological targets. We release the model, code, a full-body X-ray segmentation dataset, and a local, easy-to-use browser-based tool at \url{https://flexray.csail.mit.edu}.
\end{abstract}

\vspace{1pt}
\noindent\rule{\textwidth}{0.5pt}
\vspace{8pt}
\end{@twocolumnfalse}
]

\begin{figure*}[!htp]
    \centering
    \includegraphics[
        width=\textwidth,
    ]{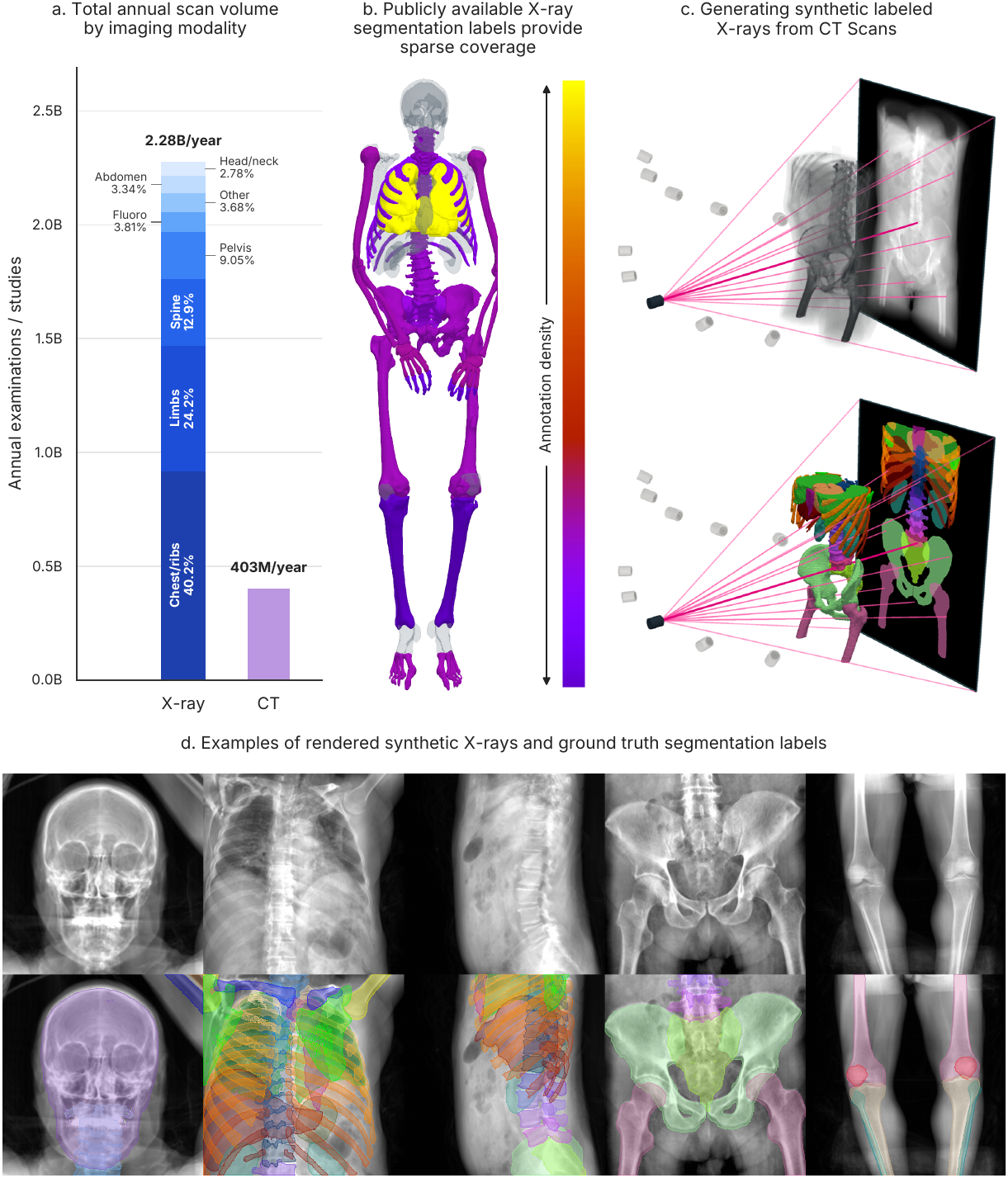}
    \caption{
    \textbf{Labeled 3D CTs can provide dense 2D X-ray supervision beyond the coverage of public X-ray segmentation datasets.}
    \textbf{a,} Global annual X-ray and CT examination volumes reported by the United Nations Scientific Committee on the Effects of Atomic Radiation~\cite{UNSCEAR2022MedicalExposure}, with X-rays subdivided by examination type.
    \textbf{b,} Despite their abundance, the labels in X-ray segmentation datasets unevenly distributed across the body and focused on the chest. Colors summarize label availability across 12 public datasets containing 667,380 labeled X-rays (\cref{sec:appendix-coverage-provenance}).
    \textbf{c,} We generate labeled synthetic X-rays by doing ray-projection through the CT volumes and their labels from different viewing angles. 
    \textbf{d,} A single labeled CT therefore generates multiple annotated training views; examples show synthetic radiographs across the body (top) and their overlapping anatomical labels (bottom).
    }
    \label{fig:overview}
\end{figure*}

\noindent Modern segmentation tools automatically outline dozens of anatomical structures in CT and MRI volumes~\cite{wasserthal2023totalsegmentator, akinci2025totalsegmentator, ferrara2026sharing, mckay2024musclemap, mann2025evaluating, billot2023synthseg}, enabling treatment planning~\cite{bibault2024deep, elguindi2019deep, pang2025multicentre}, population-scale studies~\cite{fumagalli2024automated, thiriveedhi2024cloud}, and routine extraction of quantitative biomarkers~\cite{fischl2012freesurfer}. However, despite their widespread clinical use and greater abundance (\cref{fig:overview}a), X-rays lack comparable tools for quantitative anatomical analyses. 

The main bottleneck is not a shortage of X-rays, but rather the difficulty of annotating them. Because X-rays collapse 3D anatomy onto a 2D detector, organs overlap and obscure boundaries that are easily separated in volumetric modalities. Creating dense pixel-level annotation of every organ for training X-ray segmentation systems is therefore difficult even for experts~\cite{seibold2022reference,lenga2018deep}. These challenges are compounded by the unique heterogeneity of X-ray: anatomical regions, patient positions, fields of view, and acquisition setups vary substantially across medical contexts~\cite{kelly2012chest, tafti2020x}, while patient- and procedure-specific factors like bone density, contrast agents, and implanted devices add further diversity and difficulty~\cite{federle2017contrast,guerri2018quantitative,mathew2019chest}. As a result, existing datasets typically annotate only a handful of organs (e.g., the heart and lungs) under restricted acquisition settings and in specific populations~\cite{gaggion2024chexmask, danilov2022chest} (\cref{fig:overview}b), leaving insufficient training data for reliable, multi-organ anatomical segmentation across the breadth of real-world X-rays. 

Overcoming this limitation would create substantial biomedical and scientific opportunities. X-rays are acquired routinely and at enormous scale, yet much of their anatomical information is still assessed only through visual interpretation or coarse manual measurements. Instead, a generalist X-ray segmentor could make radiography \textit{quantitative} by converting each X-ray into a structured anatomical map, enabling automatic and consistent extraction of anatomical measurements, imaging biomarkers, and longitudinal changes. Further, retrospective archives and biobanks could be mined to determine population-level associations between anatomy, disease, treatment, and outcomes using a modality that is already routinely acquired worldwide.

We address this bottleneck with a scalable, physics-based generative data engine and use it to train \textbf{\method{}}, a generalist model that automatically segments 60 anatomical structures in clinical X-rays. Instead of relying on painstaking manual annotation, our data engine uses several densely labeled 3D CT datasets to simulate X-rays (digitally reconstructed radiographs or DRRs~\cite{unberath2018deepdrr, gopalakrishnan2022fast}) together with dense anatomical labels (\cref{fig:overview}c,d), providing broad whole-body coverage across age groups. Because the data engine is fully controllable, we also randomize pose, projection geometry, magnification, and organ density to sample annotated training examples across a wide range of acquisition conditions~(\cref{fig:data-engine}a).

However, dense synthetic supervision alone does not guarantee generalization to real radiographs due to order-of-magnitude differences in resolution between CT and X-ray, content differences such as inscribed text, and the challenges of modeling higher-order X-ray image formation terms~\cite{gao2022synthex}. We therefore use generative image-editing models~\cite{flux-2-2025} to introduce clinical appearance factors that are difficult to model analytically, including scanner artifacts, text overlays, scatter, and other higher-order effects~(\cref{fig:data-engine}b). Further, because whole-body CT collections often omit some regions that are commonly imaged with X-rays (e.g., the hands and feet), we supplement our CT training sources with real X-ray datasets to close coverage gaps across common X-ray use cases. More broadly, we treat dataset construction as an iterative coverage problem: we identify and close gaps in the training distribution by including targeted data sources, custom X-ray-specific augmentations, and carefully tuned data mixtures. Together with our synthetic labeled X-ray generation pipeline, this combined dataset spans greater anatomical, appearance, and geometric diversity than any existing collection.

Trained using this generative data engine, \method{} accurately segments anatomy across held-out research datasets and in-the-wild X-rays, spanning regions, populations, imaging setups, and medical conditions. Beyond segmentation accuracy, we demonstrate how making X-rays quantitative enables automated measurements for disease grading, robust surgical X-ray guided navigation, and data-efficient learning of previously unseen pathological targets. We release \method{}'s trained models and data engine as open-source software, together with our generatively enhanced X-ray dataset (\url{https://huggingface.co/datasets/VictorButoi/flexray-data}) and a one-click, browser-based tool for segmenting any X-ray at \url{https://flexray.csail.mit.edu}.

\section*{Results}

\begin{figure*}[t!]
    \centering
    \includegraphics[width=\textwidth]{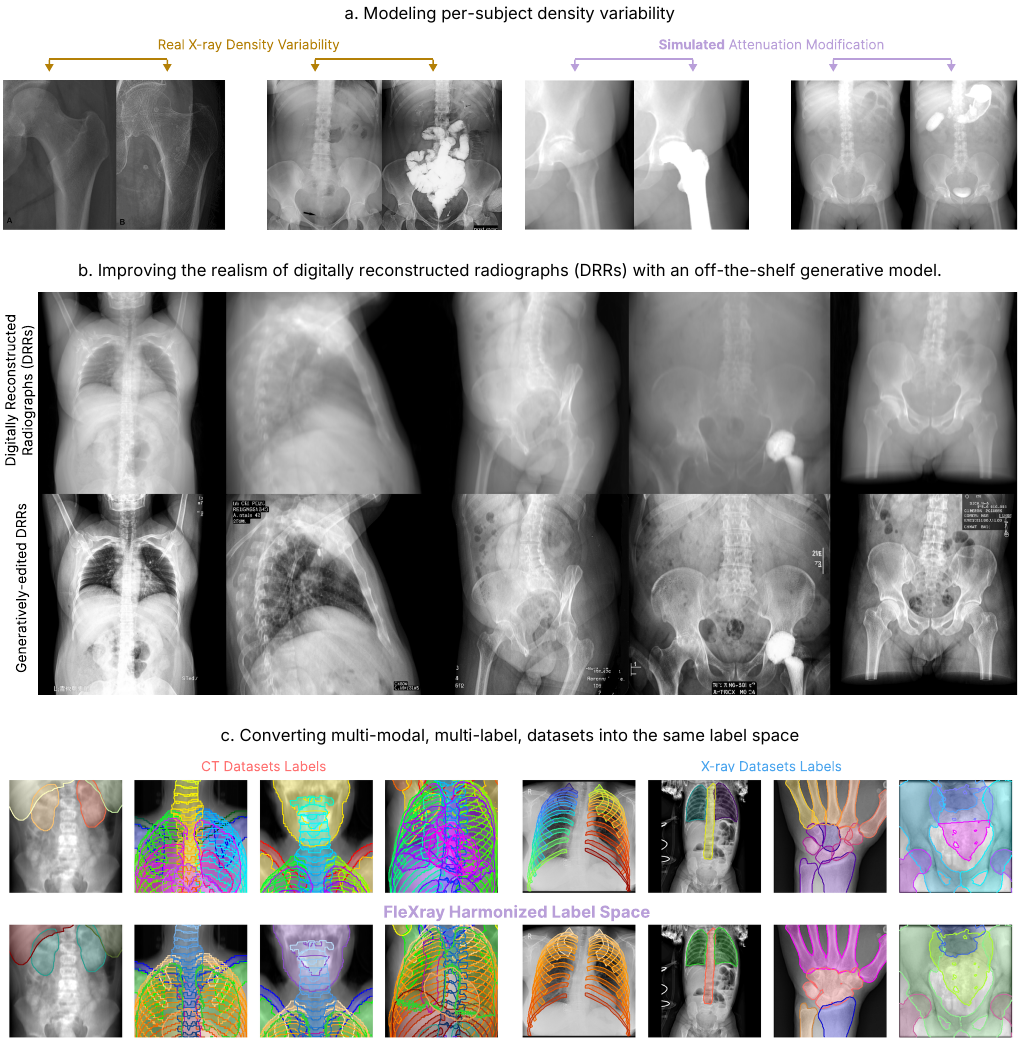}
    \caption{\textbf{Our generative data engine broadens radiographic appearance and harmonizes inter-dataset label formats.} \textbf{a,} Clinical X-ray appearance varies with factors including bone pathology~\cite{jang2021prediction} and contrast agents~\cite{yarmohammadi2010vicarious} (left). We use per-structure attenuation changes to introduce label-specific density randomization during training (right). \textbf{b,} An off-the-shelf generative image editor transforms raw digitally reconstructed radiographs (DRRs; top) into realistic X-rays with clinical textures and acquisition artifacts (bottom). \textbf{c,} Because our source datasets use different label definitions, we map their annotations (top) into a common 60-structure vocabulary (bottom), allowing CT and real X-ray supervision to be freely combined.}
    \label{fig:data-engine}
\end{figure*}

\paragraphsection{\method{} learns full-body X-ray segmentation from a generative data engine.}

\subpara{Scaling anatomical coverage across the body} Labeled X-ray simulation from CT datasets provides dense anatomical supervision beyond the limited coverage of public X-ray annotations (\cref{fig:overview}b). Prior work has rendered labeled radiographs from segmented CT volumes~\cite{unberath2018deepdrr,gao2022synthex,seibold2023accurate,alshenoudy2025leveraging,killeen2025fluorosam,dong2025anycxr}, but generally targets specific regions, particularly the chest~\cite{unberath2018deepdrr,gao2022synthex,seibold2023accurate,dong2025anycxr}, standard frontal/lateral projections~\cite{alshenoudy2025leveraging}, or interactive workflows~\cite{killeen2025fluorosam}. Our goal is instead to scale simulation across the body, patient populations, and continuously varying acquisition parameters and geometries.

We aggregate six CT datasets comprising 2,159 subjects, including 1,597 densely annotated whole-body CTs from the MOOSE dataset~\cite{ferrara2026sharing}. Because ``whole-body'' CT scans generally cover only neck-to-knee anatomy, use standardized patient poses, and rarely include pediatric subjects, we add five targeted CT datasets to fill gaps in anatomical and population coverage~\cite{podobnik2024han,cheng2021automatically,lin2023rsna,jordan2022pediatric,wang2026elbow}. However, peripheral anatomy remains sparsely imaged in public CT databases. Therefore, we incorporate four real X-ray datasets totaling 764 samples: public hand and foot annotations~\cite{HandBones,FootBones}, plus manual forearm and upper-arm annotations that we collect from MURA~\cite{rajpurkar2017mura}.

Combining these sources requires reconciling differences in anatomical coverage and label definitions between source datasets, such as ``spine'' in one dataset versus ``L1 vertebra'' in another. We therefore develop an automatic harmonization pipeline that hierarchically maps dataset-specific ontologies into a common 60-structure label space (\cref{fig:data-engine}c; \cref{sec:appendix-label-harmonization}). We restrict this vocabulary to anatomy meaningfully represented in radiographs and exclude structures generally invisible on X-ray, such as the brain and skeletal muscle.

\subpara{Sampling anatomy across X-ray acquisition conditions} We train on DRRs sampled on-the-fly during training from multiple CT sources (\cref{fig:overview}c,d; \cref{sec:appendix-dataengine-CT,tab:methods-online-drr}). A physics-based analytical X-ray renderer~\cite{gopalakrishnan2022fast} generates these images from the CT volumes and specified scanner intrinsics and extrinsics. At each training iteration, we sample an anatomical target, viewpoint, field of view, magnification, and projection geometry. We also randomize per-organ attenuation to simulate organ density variations independently of acquisition geometry (\cref{fig:data-engine}a; \cref{sec:appendix-density-mods}). Because several structures can contribute to the same detector pixel, we project every 3D anatomical mask independently. The resulting targets preserve the dense, overlapping nature of X-ray anatomy rather than imposing a mutually exclusive segmentation map (\cref{fig:overview}c,d; \cref{sec:drr-rendering}).

\subpara{Closing the synthetic-to-real appearance gap} Analytical DRRs capture anatomy, geometry, and attenuation, but do not reproduce the full appearance distribution of clinical X-rays. Resolution differences, text overlays and higher-order effects such as dosage, detector response, and scatter remain difficult to model efficiently~\cite{gao2022synthex}. We therefore apply a pretrained language-conditioned generative image editor~\cite{flux-2-2025} to DRRs sampled from the MOOSE dataset to add coverage for higher-order clinical effects (\cref{fig:data-engine}b; \cref{sec:appendix-dataengine-enhanced-drrs}). The generative editor requires no task-specific fine-tuning and is instructed to introduce clinical appearance while preserving projected anatomy. While generative models have previously been used to synthesize chest X-rays~\cite{sundaram2021gan,chambon2022roentgen,ribeiro2026scaling}, ours is the first whole-body, label-preserving generative simulation-to-real pipeline for constructing a reusable densely labeled X-ray dataset. Further, because generative editing can occasionally alter anatomy and introduce image-label disagreement, we develop an automatic quality-control pipeline that rejects simulations with gross deviations, removing approximately 4\% of enhanced DRRs (\cref{fig:appendix-flux-filtering,sec:appendix-dataengine-enhanced-drrs}).

\begin{figure*}[p]
    \centering
    \includegraphics[width=\textwidth]{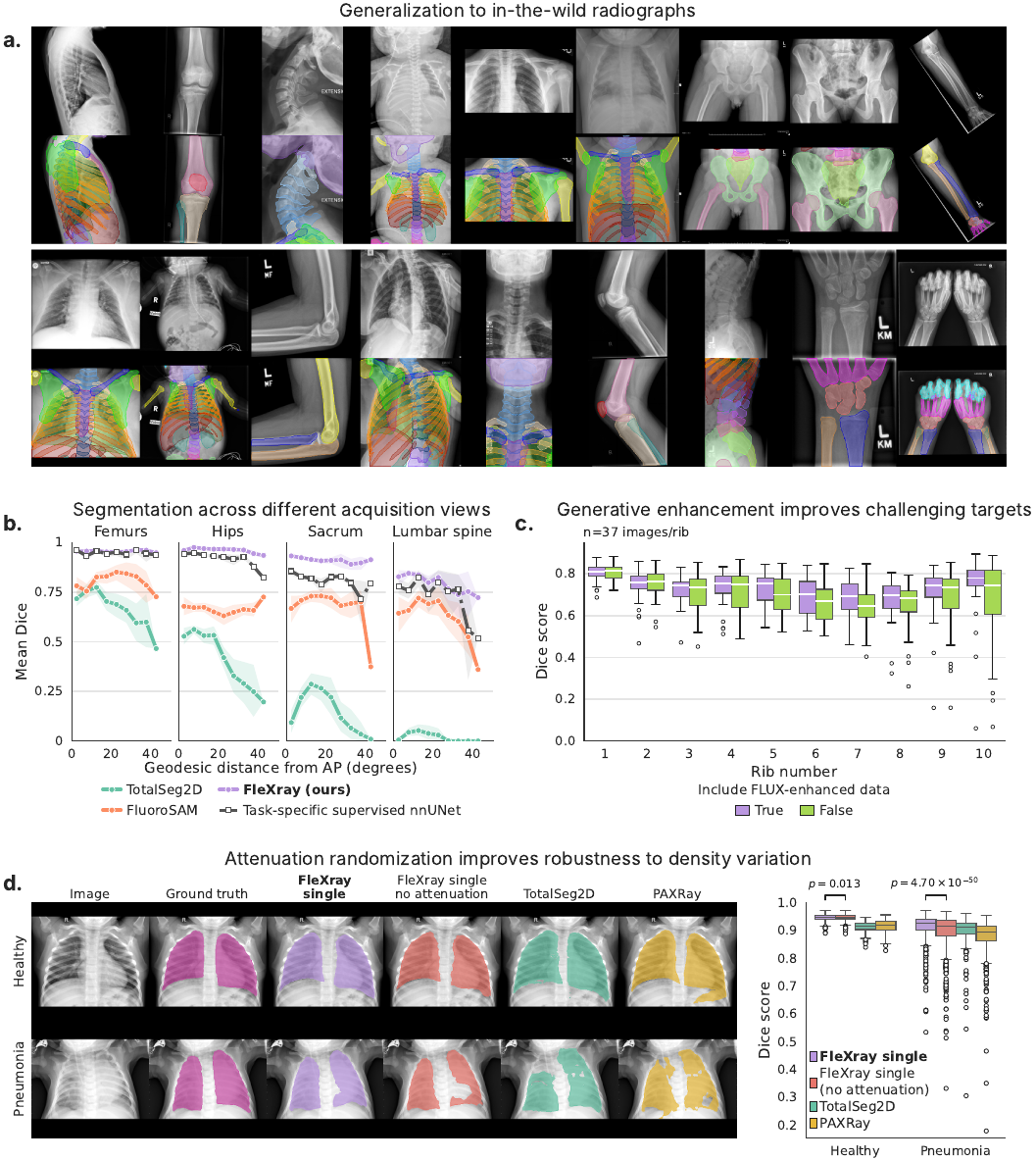}
    \caption{\textbf{\method{} generalizes across acquisition conditions.} \textbf{a,} Ensemble predictions on 18 unlabeled Radiopaedia radiographs (attributions in \cref{sec:appendix-radiopaedia-attributions}). \textbf{b,} Mean per-structure Dice versus geodesic distance from the frontal view on DeepFluoro (\textit{n}$=$213 frames from two held-out subjects), grouped into 5$^\circ$ bins. Shading shows confidence intervals obtained by resampling label--frame pairs within each bin; pairs are included only when the ground-truth label occupies at least 0.1\% of the image. \textbf{c,} Per-rib Dice on VinDr-Rib (\textit{n}$=$37 images per rib) for \method{} (single) trained with generatively edited or unedited DRRs, showing improvements for the lower ribs. \textbf{d,} Example lung segmentations (left) and per-image Dice (right) on DarwinCVD19, comparing \method{} (single) with and without attenuation randomization against TotalSegmentator2D and PAXray. Results are stratified into healthy (\textit{n}$=$241) and pneumonia (\textit{n}$=$711) cases; brackets report \textit{p} values between the two \method{} variants.}
    \label{fig:robustness-eval}
\end{figure*}

The final data engine combines three complementary data sources: online analytic projections for anatomical and geometric diversity, quality-controlled generatively-edited DRRs for realistic appearance, and targeted real X-rays for regions poorly represented in CT (\cref{tab:methods-training-data}). The utility of the online/offline mixture and the contribution of generative editing are examined separately in \cref{tab:ablation-enhanced-drrs-proportion} and \cref{fig:robustness-eval}c. We release the resulting dataset, including the quality-controlled enhanced DRRs and our MURA annotations, as part of our data release (\cref{sec:appendix-data-sources}).

\subpara{Training on incomplete and overlapping annotations} We train \method{} on our data engine's outputs. Combining heterogeneous datasets increases coverage but means that each source annotates only a subset of the 60-structure vocabulary. Treating unannotated structures in a source as \textit{absent} would therefore mislead training. Instead, we route each source through a loss matched to its annotation coverage, ignoring labels whose status is undetermined. Where absence is anatomically certain, we include impossible structures as explicit negatives; for example, hand structures can safely be supervised as absent in foot X-rays. This dataset-routed objective allows complementary sources to be learned jointly without assuming that unannotated anatomy is absent (\cref{sec:appendix-training-objective}). Further, X-ray segmentation is intrinsically multi-label at the pixel level due to organ overlap. We therefore derive supervision directly from the projection geometry, mark every organ intersected by a ray as positive, and predict an independent probability map for each structure (\cref{fig:overview}c,d; \cref{sec:appendix-network-architecture,sec:appendix-training-objective}).

The final \method{} system is an ensemble of five independently trained networks exposed to different proportions of generatively enhanced data, with each network using 16-sample test-time augmentation (\cref{sec:appendix-network-architecture,sec:appendix-tta}). Ensembling and test-time-augmentation aim to broaden the data distribution at training and inference, respectively, both seeking to improve robustness on in-the-wild X-rays (\cref{fig:appendix-tta-sweep}). For the ablations and finetuning experiments that follow, we use a single model---\textbf{\method{} (single)}---trained with equal proportions of standard and generatively enhanced X-rays. Together, this pipeline converts heterogeneous CT and X-ray datasets into a harmonized source of dense supervision spanning anatomy, populations, projection geometries, and clinical appearance, enabling the training of full-body X-ray segmentation networks that require no user prompts or acquisition metadata at inference.

\begin{figure*}[p]
    \centering
    \includegraphics[width=\textwidth]{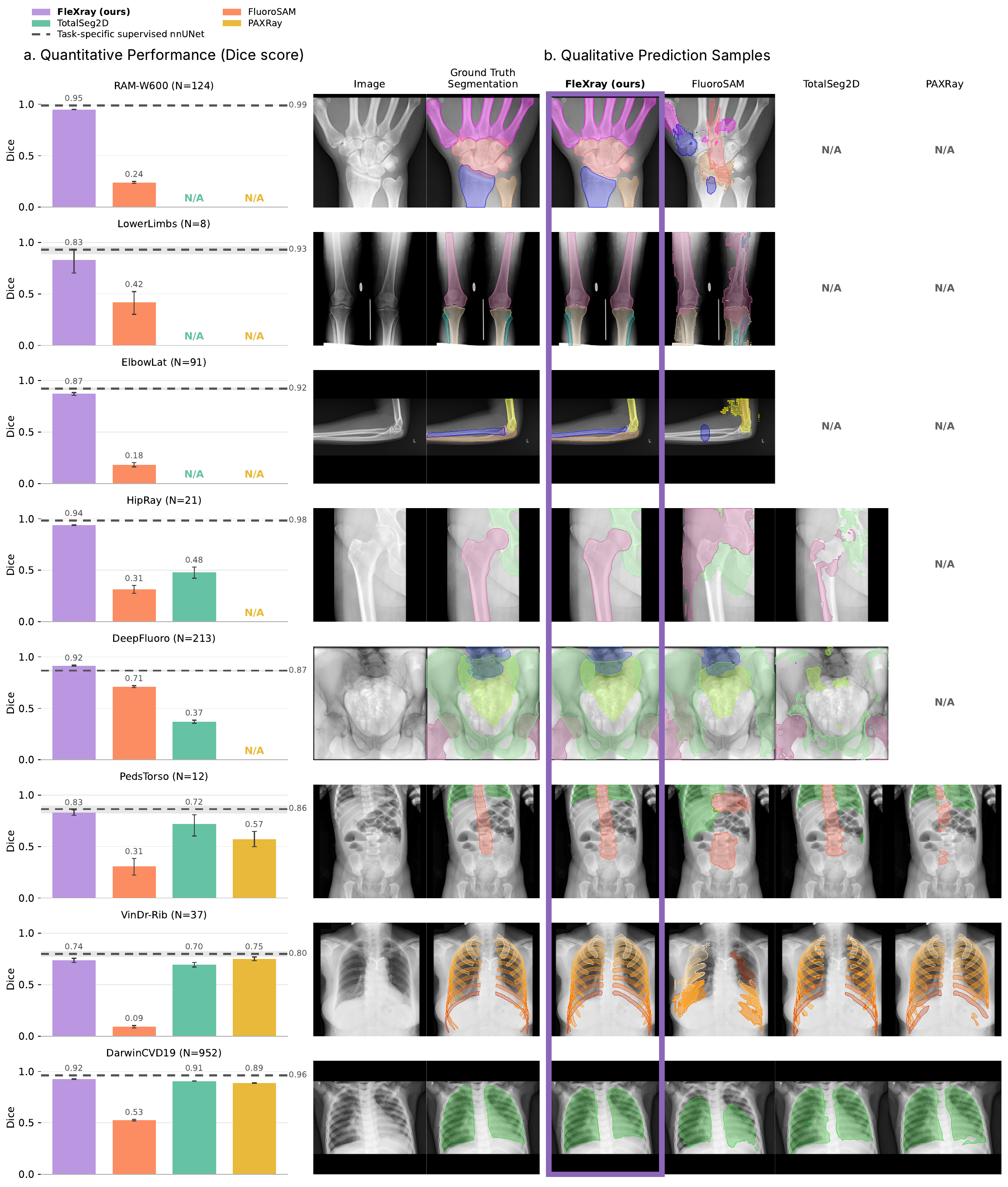}
    \caption{\textbf{\method{} generalizes across eight held-out X-ray datasets.} Each row shows a dataset entirely withheld from \method{} training; $N$ denotes the number of test images. \textbf{a,} Mean label-averaged Dice across test images for \method{}, PAXray, TotalSegmentator2D, and FluoroSAM, with 95\% confidence intervals. FluoroSAM receives a ground-truth-derived text prompt and positive click for each target. Dataset-specific nnU-Nets trained on the corresponding training splits provide in-domain supervised references (dashed lines and shaded confidence intervals). N/A indicates that a method does not predict the labels required for evaluation on that dataset. Full Dice results are in \cref{tab:appendix-baseline-dice}. \textbf{b,} Example radiographs, ground-truth labels, and prediction overlays thresholded at 0.5. The \method{} column is outlined in purple.}
    \label{fig:quant-eval}
\end{figure*} 

\paragraphsection{\method{} robustly segments real X-rays across medical contexts.}

\subpara{Quantitative segmentation evaluation} 
We evaluate \method{} on eight held-out X-ray datasets spanning the limbs, pelvis, spine, and chest, spanning multiple fields-of-view and populations (\cref{fig:quant-eval}; \cref{tab:methods-eval-data}). Where possible, we compare against existing X-ray segmentation frameworks that aim to generalize to the target dataset, including PAXray~\cite{seibold2023accurate}, TotalSegmentator2D~\cite{alshenoudy2025leveraging}, and FluoroSAM~\cite{killeen2025fluorosam}. PAXray and TotalSegmentator2D are evaluated only on supported anatomical regions, whereas FluoroSAM is evaluated throughout but receives a simulated ground-truth-derived text prompt and positive click for each target. As in-domain supervised upper bounds, we also train a separate nnU-Net on the training split of each evaluation dataset (\cref{sec:appendix-baselines}). Predictions are post-processed to match dataset-specific annotation conventions (\cref{sec:appendix-postprocessing}) and evaluated using Dice score (\cref{sec:appendix-metrics}).

Across all eight datasets, \method{} achieves an equal-dataset mean Dice of 0.875 and significantly outperforms every supported baseline on seven datasets (\cref{fig:quant-eval}, left; all \textit{p}$<$0.001 on six datasets). On VinDr-Rib, its performance is not significantly different from PAXray (\textit{p}=0.146), a baseline specifically developed for chest X-rays. Qualitatively, \method{} remains anatomically coherent despite large differences in field of view, projection angle, image polarity, acquisition modality, and patient population (\cref{fig:quant-eval}, right).

Although \method{} never trains on any of these evaluation datasets, its performance is comparable to that of supervised nnU-Nets trained for that specific dataset. \method{} trails these in-domain networks by only 0.039 Dice on average (0.875 versus 0.914), and on DeepFluoro it performs \textit{better} than the nnU-Net trained directly on the dataset (0.915 versus 0.866). This is due to DeepFluoro containing little training data and spanning a wide range of acquisition angles, illustrating the advantages of learning from a broad synthetic acquisition distribution.

However, while held-out, academic datasets still represent relatively curated imaging conditions. We therefore also apply \method{} to ``in-the-wild'' radiographs collected from Radiopaedia (\cref{fig:robustness-eval}a). These examples span frontal, lateral, and oblique views; unilateral and bilateral examinations; adult and pediatric anatomy; and large differences in field of view. Across these heterogeneous cases, \method{} continues to produce qualitatively coherent anatomical segmentations, providing evidence that its robustness extends beyond benchmark datasets.

\subpara{Robustness to acquisition angle}
Clinical radiographs are frequently acquired at oblique angles for certain examinations, whereas most X-ray segmentation datasets and models focus entirely on standard frontal and lateral views~\cite{singh2018deep}. We isolate this factor using DeepFluoro, which provides both labels and ground-truth acquisition angles for each X-ray.
Across the 213 frames from its two held-out test subjects, we measure per-structure Dice as a function of geodesic distance from the frontal view (\cref{fig:robustness-eval}b). Overall, \method{} achieves a mean Dice of 0.915 [0.912, 0.919], with 5$^\circ$-bin averages ranging from 0.925 [0.921, 0.928] at 5--10$^\circ$ to 0.888 [0.847, 0.923] at 35--40$^\circ$. Femurs, hips, and sacrum remain within 0.05 Dice of their frontal-view performance across the full pose range. Only the lumbar spine, which becomes increasingly small and foreshortened in oblique views, shows a larger decline, from 0.840 [0.829, 0.850] within 15$^\circ$ of frontal (\textit{n}$=$136) to 0.751 [0.690, 0.805] beyond 25$^\circ$ (\textit{n}$=$27). In comparison, the applicable baselines (FluoroSAM and TotalSegmentator2D) are both less accurate and less stable, reaching mean Dice scores of 0.711 [0.701, 0.722] and 0.371 [0.357, 0.386], respectively. Thus, \method{} remains robust well beyond the standard acquisition angles represented in most X-ray datasets.

\subpara{Ablating generative enhancement}
To isolate whether generative X-ray enhancement provides useful supervision in addition to our augmentation pipeline, we remove it as an augmentation. While the equal-dataset average Dice is unchanged (0.870 enhanced versus 0.870 raw; \textit{n}$=$1,458), we find substantial benefits in regions that are challenging to render faithfully in rendered DRRs. For example, as the lower ribs on VinDr-Rib are obfuscated by abdominal organs in standard DRRs (\cref{fig:robustness-eval}c), ribs 5--10 all improve with generative enhancement (+3.9 [+2.2, +5.8] pooled, \textit{p}$<$0.001), with gains increasing from +2.5 Dice points at rib 5 to +7.4 at rib 10. 
These results suggest that generative editing contributes useful appearance variation beyond conventional augmentation, with substantial benefit to specific anatomical structures.

\subpara{Ablating attenuation randomization} We now test whether per-label attenuation randomization (\cref{sec:appendix-density-mods}) improves robustness. Averaged across datasets, the effect is modest as equal-dataset mean Dice increases only from 0.863 to 0.870 with randomization. However, in pathological datasets, such as DarwinCVD19, we find that lung Dice increases from 0.907 to 0.922 (\textit{n}$=$952, \textit{p}$<$0.001). Stratifying by diagnosis (\cref{fig:robustness-eval}d) shows that this gain is negligible in healthy lungs (0.945 to 0.946, \textit{n}$=$241), but concentrated in pneumonia cases where the lungs are dense and Dice rises from 0.894 to 0.914 (\textit{n}$=$711, \textit{p}$<$0.001; Welch \textit{t}-test between groups, \textit{p}$<$0.001). Correspondingly, the number of pneumonia images below 0.85 Dice falls from 122 to 50. Therefore, as intentionally designed, attenuation randomization improves robustness when pathology substantially alters radiographic density.

\begin{figure*}[t!]
    \centering
    \includegraphics[width=\textwidth]{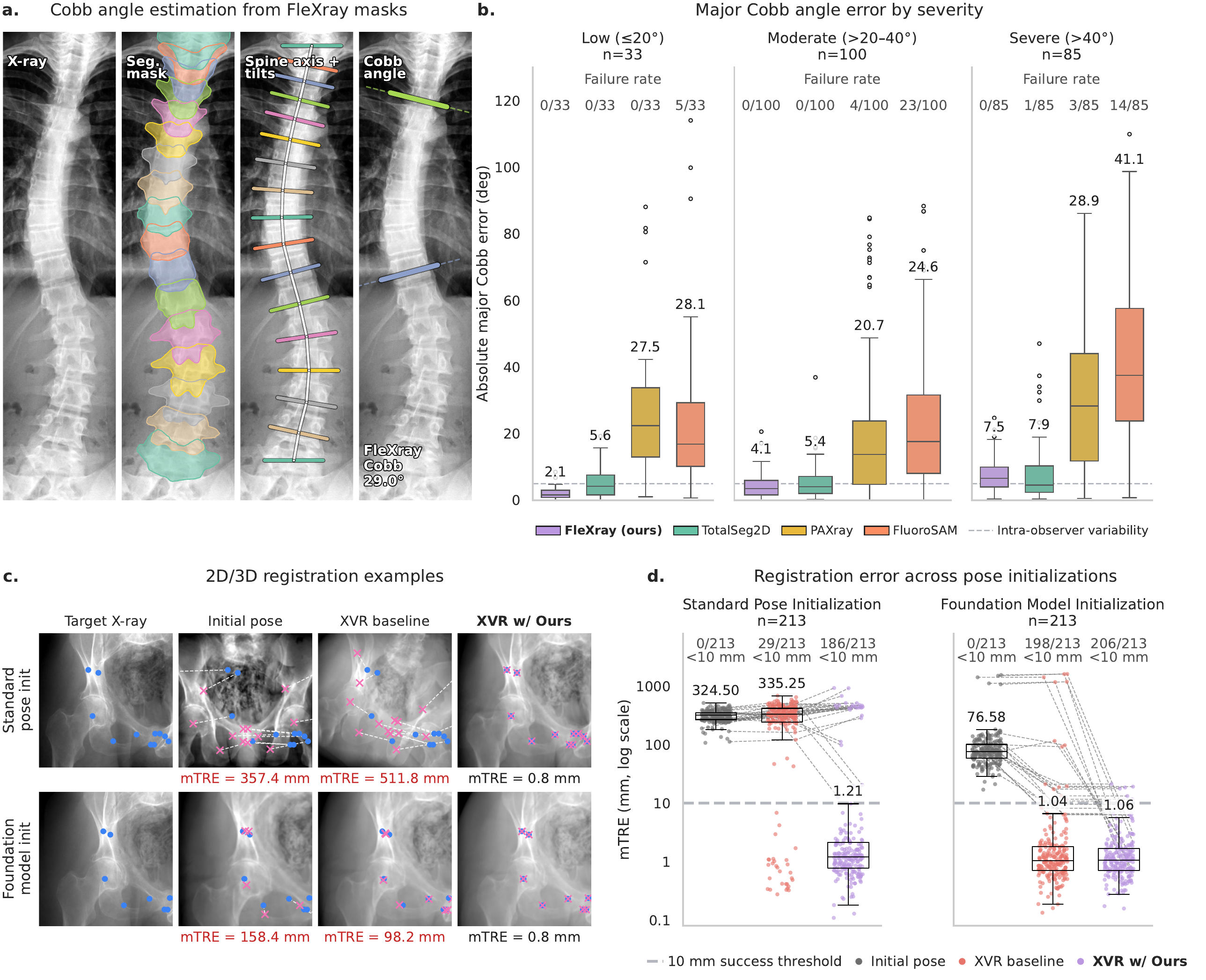}
    \caption{\textbf{\method{} enables downstream clinical tasks.} \textbf{a,} From an AASCE X-ray, \method{} predicts vertebra masks, from which centerline-derived end-plate orientations yield the major Cobb angle. \textbf{b,} Absolute major Cobb-angle error by reference severity, over the images on which each method yields a measurable curve. Counts at the top give the number of unmeasurable images (failed/total), which are excluded from that method's distribution. Mean errors are shown above the whiskers. \textbf{c,} Visualizations of the C-arm poses estimated by \xvr and \xvr with our \method{}-derived Chamfer loss term compared to the ground truth C-arm poses from the test split of the DeepFluoro dataset with two pose initializations: a manual frontal initial pose (\textit{top}; \cite{grupp2020automatic}) and a foundation model initialization (\textit{bottom}; \cite{gopalakrishnan2026rapid}). Poses with mTRE $>$10~mm are shown in red. \textbf{d,} Mean target registration error (mTRE) for each X-ray for each initialization strategy and registration method. The dashed line represents the \SI{10}{mm} success threshold. Boxes show the median, IQR, and Tukey whiskers at 1.5$\times$IQR; all \textit{n}$=$213 X-rays are shown as points. Gray trajectories denote the same sample evaluated by different methods.}
    \label{fig:real-world-eval-ootb}
\end{figure*}

\paragraphsection{\method{} unlocks diverse downstream clinical and scientific workflows.}
\noindent We next evaluate whether \method{} is useful beyond segmentation itself. We consider three complementary modes of reuse: directly converting its anatomical masks into clinical measurements, using them as semantic priors within downstream workflows, and adapting it for new anatomical and pathological targets.

\subpara{\method{} predictions quantify scoliosis severity within inter-rater variability} Scoliosis severity is routinely quantified from X-rays using the Cobb angle, a standard measure of spinal curvature~\cite{langensiepen2013measuring,cobb1948outline}. However, manual measurements vary by approximately 3--5$^\circ$ within observers and 6--7$^\circ$ between observers~\cite{morrissy1990measurement,negrini2016sosort}. We therefore ask whether \method{} can recover this measurement directly from its vertebral segmentations on the Accurate Automated Spinal Curvature Estimation (AASCE) MICCAI 2019 challenge dataset~\cite{wang2021evaluation}, which provides reference Cobb-angle annotations on anterior-posterior spinal X-rays. Following clinical guidelines~\cite{negrini2016sosort}, we stratify curves as low ($\leq$20$^\circ$), moderate (20--40$^\circ$), or severe ($>$40$^\circ$). Predictions with fewer than three usable vertebrae are reported separately as failures and the segmentation to Cobb angle conversion is detailed in~\cref{fig:real-world-eval-ootb}a and \cref{sec:appendix-cobb}.

On AASCE, \method{} produces a measurable curve with no failures for all 218 images and achieves the lowest overall error, with a mean absolute error (MAE) of 5.13$^\circ$ [4.55, 5.74] (\cref{fig:real-world-eval-ootb}b). TotalSegmentator2D fails on 1 image and reaches 6.38$^\circ$ [5.51, 7.32] MAE on the remainder, compared with 24.94$^\circ$ [22.06, 27.91] for PAXray, which fails on 7 images, and 31.82$^\circ$ [28.19, 35.58] for FluoroSAM, which fails on 42. \method{} also attains the lowest MAE in every severity stratum (2.08$^\circ$, 4.12$^\circ$, and 7.50$^\circ$ for low, moderate, and severe curves), significantly so in every comparison (\textit{p}$\le$0.024), except against TotalSegmentator2D for severe curves (7.87$^\circ$; \textit{p}$=$0.70).

Importantly, \method{}'s error lies below the 6--7$^\circ$ variability between expert observers. For moderate curves, which are particularly relevant to clinical planning~\cite{richards2005standardization}, its 4.12$^\circ$ MAE also falls within the 3--5$^\circ$ variability of repeated measurements by a single expert. Overall, 61\% [55, 67] of estimates fall within 5$^\circ$ of the reference and 88\% [83, 92] within 10$^\circ$. Thus, general-purpose anatomical segmentations derived from \method{} can be used to power application-specific quantitative measurements and grading workflows.

\subpara{\method{} segmentations improve the robustness of 2D/3D registration}
Many image-guided procedures use preoperative 3D imaging for planning but rely on live 2D X-ray fluoroscopy during intervention for navigation~\cite{abumoussa2023machine, naik2022hybrid, metz2009patient, wagner20164d, huynh2020artificial, kim2022telerobotic}. Aligning these images through 2D/3D registration can recover 3D anatomical context during navigation~\cite{unberath2021impact}. Currently, conventional methods optimize image similarity between a real intraoperative X-ray and a DRR rendered from a candidate 3D pose from the preoperative CT~\cite{penney1998comparison, knaan2003effective, grupp2020automatic, grupp2019pose, bier2019learning, shrestha2024rayemb, gopalakrishnan2026rapid}. However, many of the same challenges with poor soft-tissue contrast and overlapping anatomy make this registration objective highly non-convex, leaving iterative registration highly susceptible to local minima without proper initialization~\cite{gu2020extended, gopalakrishnan2022fast, gao2023fully}.

We therefore use \method{}'s anatomical predictions as auxiliary semantic supervision for registration. Specifically, we add a bidirectional Chamfer loss~\cite{tenenbaum1977parametric, borgefors1986distance} that aligns distance transforms of structures segmented in the real X-ray with projections of the corresponding structures from the preoperative CT (\cref{sec:appendix-registration}). We incorporate this loss into \xvr, an image-based iterative X-ray to volume registration pipeline~\cite{gopalakrishnan2026rapid}. Unlike image similarity, whose useful gradient can disappear when the DRR and X-ray have little overlap, the segmentation-based Chamfer term remains informative from substantially worse initial poses (\cref{sec:appendix-registration}).

We evaluate on all 213 held-out DeepFluoro X-rays with ground-truth C-arm poses starting from two possible initialization strategies (\cref{fig:real-world-eval-ootb}c). Performance is measured using mean target registration error (mTRE) (\cref{sec:appendix-metrics}) and gross failure rate (GFR), defined as the fraction of registrations with mTRE greater than \SI{10}{mm}. From a manually selected frontal pose, a common clinical initialization~\cite{grupp2020automatic}, the median starting error is 324.5~mm with 100\% GFR. Image-based optimization alone remains trapped at a median mTRE of 335.3~mm (86.4\% GFR). However, adding the \method{}-enabled Chamfer loss reduces median mTRE to 1.2~mm (\textit{p}$<$0.001) and GFR to 12.7\% (\cref{fig:real-world-eval-ootb}d).

To measure potential benefits over the state-of-the-art, we next initialize optimization using a foundation model trained to predict C-arm pose directly from an X-ray~\cite{gopalakrishnan2026rapid}, yielding a median initial mTRE of 76.6~mm (100\% GFR). Image-based refinement already reduces GFR to 7.0\%, but leaves eight catastrophic failures with errors between 93 and 1,628~mm. Adding the Chamfer term recovers all eight cases with mTRE $>$50~mm and brings each below 10~mm, reducing overall GFR to 3.3\% and worst-case mTRE to 21~mm. This additional semantic loss does not sacrifice accuracy on cases that already register successfully: among the 198 frames solved by both methods, median mTRE remains 1.01 versus 1.04~mm (\cref{fig:real-world-eval-ootb}d). Thus, \method{} automatically injects useful semantic information into existing 2D/3D registration methods, widening the capture range of conventional optimization while suppressing catastrophic failures from learned initializations.

\begin{figure*}[p]
    \centering
    \includegraphics[width=\textwidth]{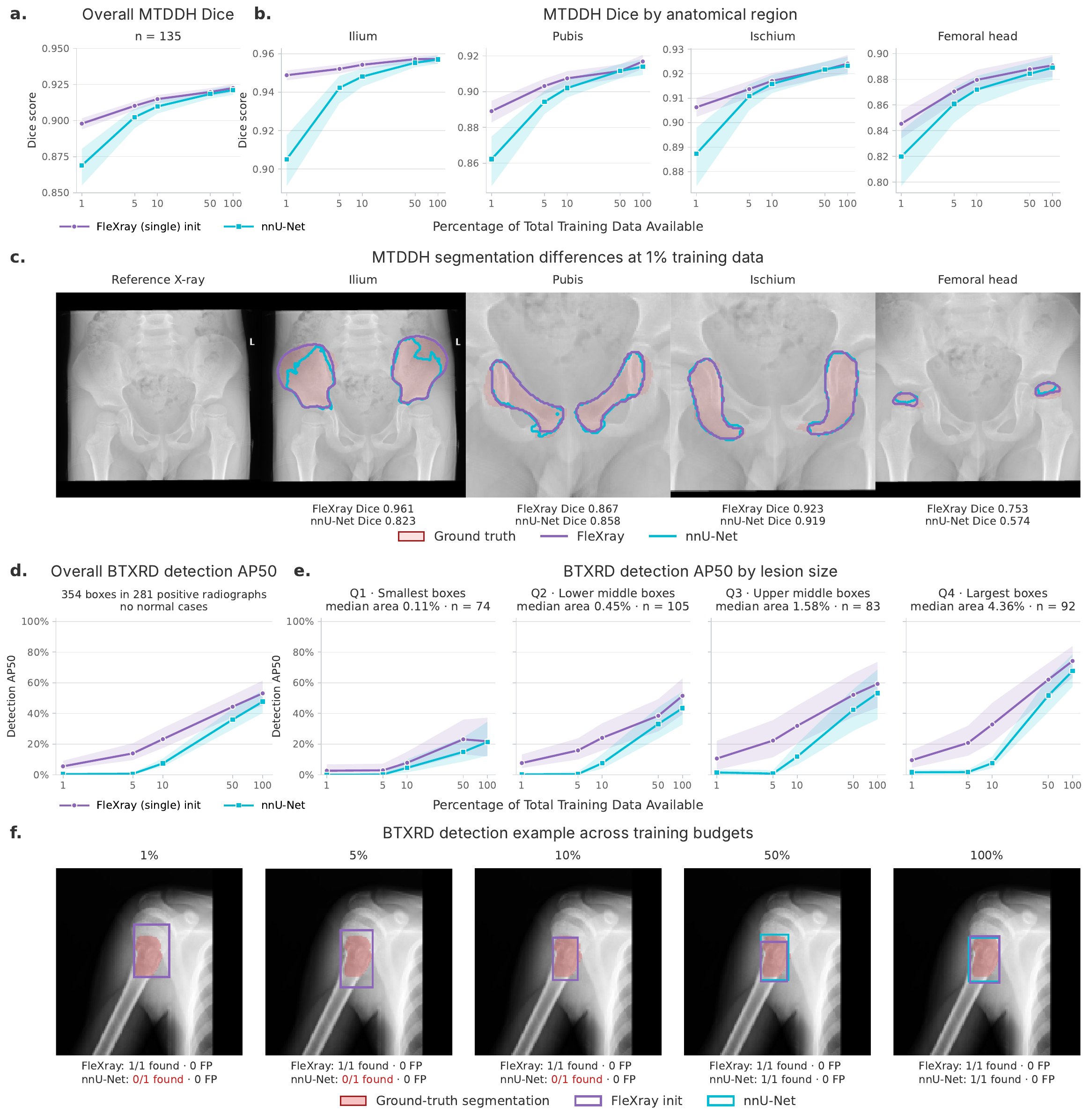}
    \caption{\textbf{\method{} provides a data-efficient initialization for new anatomical and pathological targets.} We fine-tune \method{} (single) and train a well tuned baseline (nnU-Net~\cite{isensee2021nnu}) from random initialization using 1\%, 5\%, 10\%, 50\%, or 100\% of each training split. \textbf{a,} Mean Dice across four MTDDH structures (\textit{n}$=$135 test images) as a function of the fraction of 632 labeled training patients. \textbf{b,} Corresponding per-structure Dice. \textbf{c,} MTDDH segmentation differences at 1\% training data for one test radiograph. Gray shows shared predictions, purple and teal show foreground predicted only by \method{} and nnU-Net, respectively, and dashed yellow contours show ground truth. Dice scores use the full image. \textbf{d,} Detection average precision at an intersection-over-union of 0.5 (AP50) on BTXRD, using fractions of 1,305 training images. The test set contains 354 tumors in 281 tumor-positive radiographs and no normal cases. \textbf{e,} BTXRD AP50 stratified by ground-truth bounding-box area. Quartile boundaries are defined on the training split, yielding 74, 105, 83, and 92 test tumors in successive size groups. \textbf{f,} Detections across annotation budgets for one test tumor from the third size quartile. Solid purple and teal boxes show fine-tuned \method{} and nnU-Net predictions on a shared crop; dashed yellow boxes show ground truth.}
    \label{fig:real-world-eval-finetune}
\end{figure*}

\subpara{\method{} adapts to unseen segmentation tasks with limited annotations} %
Finally, we ask whether \method{} provides a useful initialization for targets that were never part of its pretraining vocabulary. We study adaptation to fine-grained pelvic anatomy in pediatric hip dysplasia (MTDDH~\cite{qi2025mtddh}) and bone-tumor detection (BTXRD~\cite{yao2025radiograph}). 
Because accurate anatomical delineation supports hip-dysplasia measurements~\cite{tonnis1976normal}, we evaluate MTDDH using Dice. For BTXRD, where the primary focus is detecting individual tumors, we use detection average precision at an intersection-over-union of 0.5 (AP50; \cref{sec:appendix-finetune}). We replace the output head of \method{} (single) with a randomly initialized task-specific segmentation head and fine-tune the full network on 1\%, 5\%, 10\%, 50\%, or 100\% of each training split. We compare against nnU-Net~\cite{isensee2021nnu} trained from random initialization (\cref{sec:appendix-finetune}), representing a well tuned segmentation baseline.

The benefit of initialization from \method{} is largest when labels are scarce (\cref{fig:real-world-eval-finetune}). On MTDDH, using only 1\% of the training set (6 patients), finetuned \method{} reaches a mean Dice of 0.896 [0.892, 0.900] versus 0.867 [0.852, 0.879] for nnU-Net, with the largest improvement on the ilium (0.948 [0.946, 0.951] versus 0.907 [0.893, 0.920]). As supervision increases, the gap asymptotically narrows, a finding consistent with the few-shot segmentation literature~\cite{dey2025learning}. The harder BTXRD task shows a larger and more persistent advantage. With 10\% of the training set ($\sim$130 cases), finetuned \method{} reaches 23\% AP50 versus 7\% for nnU-Net, a difference of 16 percentage points [11, 22]. With only 1\% (13 cases), performance is 5\% versus 0.4\%, and nnU-Net does not match \method{}'s 5\%-budget AP50 until it receives 50\% of the training set (\cref{fig:real-world-eval-finetune}d).

\section*{Discussion}

\paragraph{Scaling generalist X-ray segmentation beyond manual annotations.}
Across all held-out segmentation datasets, \method{} matched or exceeded prior methods without requiring label-set-specific models (as in TotalSegmentator2D~\cite{alshenoudy2025leveraging}) or per-structure prompts (FluoroSAM~\cite{killeen2025fluorosam}), with no method being a consistent second place. \method's performance also approached that of dataset-specific custom networks trained directly on the evaluation domains. These results suggest that successful synthetic training depends less on reproducing any single notion of realism than on achieving broad coverage in training of the anatomical, geometric, and appearance variation encountered at deployment. For example, physics-based projection alone transfers poorly to real radiographs, while appearance randomization accounts for most of the gain (\cref{sec:appendix-ablation-augmentation}). Generative X-ray enhancement provides additional benefit where analytic rendering is least faithful (\cref{fig:robustness-eval}c), whereas real X-ray datasets of the limbs (constituting only 4\% of training iterations) close gaps in the coverage offered by public CT datasets (\cref{sec:appendix-ablation-training-datasets}).

\paragraph{Making X-rays a quantitative modality.}
The broader value of \method{} is not just the segmentation masks themselves, but the quantitative workflows they make possible. For example, its vertebral predictions disease grading measurements to within inter-rater variability. In 2D/3D registration, the same anatomical predictions provide semantic gradients where image similarity becomes uninformative, removing all catastrophic optimization failures (mTRE $>$50~mm) without degrading successful registrations. For unseen segmentation targets, \method{} provides a robust initialization that is most useful when labels are scarce. These examples underscore a broader opportunity: one anatomy can be extracted reliably from routine radiographs, X-rays can support analyses that have traditionally been confined to time consuming or ionizing volumetric modalities such as MRI or CT, respectively. Further, retrospective archives and biobanks can now be mined for anatomical biomarkers, longitudinal changes, and associations between anatomy, disease, treatment, and outcomes without requiring dedicated prospective imaging or task-specific annotation for every new analysis or region of interest.

\paragraph{A reusable resource for full-body X-ray analysis.}
We release a first-of-its-kind enhanced DRR dataset and accompanying training resources (\cref{sec:appendix-data-sources}). The collection varies anatomy, projection geometry, and appearance across the body and further includes quality-controlled generatively-enhanced DRRs, our manual forearm and humerus annotations for MURA, and seven re-distributable real X-ray sources in the format used by \method{}. We also fix the image-level splits and exclusions for every evaluation dataset, allowing future methods to be compared on identical held-out samples. Because our results indicate that performance depends strongly on the training distribution, we expect this resource to lower the barrier to the development of pan-anatomy X-ray models beyond the architecture studied here.

\paragraph{Relation to prior work.}

Prior DRR-based segmentation methods use a single projection model, either parallel~\cite{seibold2023accurate} or cone-beam~\cite{dong2025anycxr,killeen2025fluorosam}, and, to our knowledge, train on a finite set of DRRs rendered before training. \method{} combines offline-generated DRRs with online rendering that varies projection models and acquisition parameters during training. Our attenuation randomization also differs from AnyCXR~\cite{dong2025anycxr}, which shares scaling factors across soft-tissue structures and independently perturbs only individual vertebrae and ribs. We instead draw an independent multiplier for every labeled structure, increasing the diversity of relative contrast. Whereas prior generative enhancement uses image translation trained on unpaired real radiographs of the target anatomy~\cite{gao2022synthex}, we apply a pretrained, general-purpose image editor~\cite{flux-2-2025} with no X-ray-specific training. Beyond these methodological differences, prior evaluations on real X-rays have been confined to one or two body regions, chiefly the chest~\cite{seibold2023accurate,dong2025anycxr,killeen2025fluorosam} and pelvis~\cite{gao2022synthex}. To our knowledge, \method{} is the first automatic model evaluated and performant on real X-rays across the body, spanning eight held-out datasets of the limbs, pelvis, spine, and chest.

\paragraph{Limitations and future work.}
As X-rays were hard to annotate before the development of our model, our evaluation is limited by a low number of evaluation datasets with available ground truth. For example, none of our quantitative experiments include X-ray datasets with substantial clinical text overlays, despite their prevalence in real X-rays. Further, the data engine inherits biases from its CT sources, which image predominantly adults in supine poses with extremities out of the field-of-view. Targeted real X-rays partly compensate for these gaps, but \method{} can still fail on unseen acquisition geometries. Importantly, pathologies, implants, and surgical hardware are also not explicitly synthesized beyond those that are already present in the source CTs, leading to failures on severe trauma and other strongly out-of-distribution cases (\cref{fig:appendix-failures}). These limitations suggest a direct path forward: remaining coverage gaps can be addressed with targeted real or synthesized examples, while our finetuning experiments show that \method{} can adapt to new pathological targets using relatively few labeled radiographs.

\paragraph{Conclusion.} 
\method{} demonstrates that broad X-ray segmentation can be learned largely from synthetic supervision when the training distribution is designed to cover the variation encountered in real radiographs. More importantly, the resulting predictions make routine X-rays amenable to quantitative measurement, downstream optimization, and data-efficient learning. Together with the released models, data engine, and dataset, this creates a foundation for the quantitative analysis of X-rays in clinical research, analogous to the advances enabled by general-purpose segmentation tools in CT and MRI~\cite{wasserthal2023totalsegmentator, billot2023synthseg}.

{
\small
\bibliographystyle{unsrtnat}
\bibliography{main, main-xvr}
}

\newpage

\clearpage

\appendix
\renewcommand{\thesection}{\Alph{section}}
\crefalias{section}{methods}
\crefalias{subsection}{methods}
\crefalias{subsubsection}{methods}

\section*{Methods}
\label{sec:methods}

\begingroup
\setcounter{tocdepth}{3}
\makeatletter
\footnotesize\@starttoc{toc}%
\makeatother
\setcounter{tocdepth}{3}%
\endgroup
\vspace{1em}

\section{Training data engine}
\label{sec:training-data-engine}

We train \method{} using a mixture of CT and X-ray data sampled from three data sources:
\begin{enumerate}
    \raggedright
    \item DRRs rendered online from CT scans (\cref{sec:appendix-dataengine-CT}).
    \item Diffusion-enhanced synthetic X-rays (\cref{sec:appendix-dataengine-enhanced-drrs}).
    \item Annotated real X-rays (\cref{sec:appendix-annotated-real-xrays}).
\end{enumerate}

\noindent We sample each dataset according to the mixture weights in Table~\ref{tab:methods-training-data}.

\begin{table*}[t]
\centering
\scriptsize
\caption{\textbf{Training data sources and mixture proportions.} Proportion is the probability that a batch is drawn from each data source at each training step (proportions sum to 100\%). The online MOOSE and enhanced-DRR proportions are paired to sum to 75\%, with the remaining sources contributing 25\%. Counts are CT volumes for CT sources and radiographs for X-ray sources; enhanced-DRR counts refer to source CT subjects. All splits are subject-disjoint. CT-derived sources reserve no test split because all evaluation is performed on the held-out real X-ray datasets (\cref{tab:methods-eval-data}). Enhanced DRRs is a dataset of DRRs pre-rendered from MOOSE and refined with the diffusion model \texttt{flux.2-klein-9b}~\cite{flux-2-2025}.}
\label{tab:methods-training-data}
\begin{tabularx}{\textwidth}{@{}lp{0.145\textwidth}p{0.165\textwidth}p{0.12\textwidth}X@{}}
\toprule
Dataset & Proportion (\%) & Source & Train / Val / Test & Supervised labels \\
\midrule
\addlinespace[0.2em]
MOOSE~\cite{ferrara2026sharing} & \{0, 25, 37.5, 50, 75\} & CT & 1,437 / 160 / 0 & Whole-body skeleton (skull, long bones, ribs 1--12, vertebrae C1--L5, hips, sacrum) and thoracoabdominal organs (lungs, heart, liver, spleen, kidneys) \\
\lightrule
HaN-Seg~\cite{podobnik2024han} & 5.0 & CT & 37 / 5 / 0 & MOOSE label space \\
\lightrule
Shoulder-CT~\cite{cheng2021automatically} & 5.0 & CT & 16 / 2 / 0 & MOOSE label space \\
\lightrule
RSNAFrac~\cite{lin2023rsna} & 4.0 & CT & 78 / 9 / 0 & Skull, cervical and upper thoracic (T1--T7) vertebrae \\
\lightrule
PedsCT~\cite{jordan2022pediatric} & 4.0 & CT & 323 / 36 / 0 & Kidneys, liver, spleen, lungs, heart \\
\lightrule
ElbowCT~\cite{wang2026elbow} & 3.0 & CT & 49 / 7 / 0 & Humeri, ulnae \\
\midrule
HandBones~\cite{HandBones} & 1.0 & X-ray & 65 / 15 / 13 & Carpals, phalanges, metacarpals, radii, ulnae \\
\lightrule
FootBones~\cite{FootBones} & 1.0 & X-ray & 400 / 86 / 85 & Metatarsals, toes \\
\lightrule
MURA Forearm~\cite{rajpurkar2017mura} & 1.0 & X-ray & 35 / 8 / 7 & Humeri, radii, ulnae \\
\lightrule
MURA Humerus~\cite{rajpurkar2017mura} & 1.0 & X-ray & 35 / 8 / 7 & Humeri, radii, ulnae \\
\midrule
Enhanced DRRs & \{0, 25, 37.5, 50, 75\} & Diffusion-refined DRR & 1,437 / 160 / 0 & MOOSE label space \\
\bottomrule
\end{tabularx}
\end{table*}

\begin{table*}[t]
\centering
\scriptsize
\caption{\textbf{Online DRR rendering settings per training CT source.} Shared defaults apply to all datasets. Each dataset specifies its own source-to-detector distance (SDD), detector pixel spacing $\Delta$, projection mode, in-plane rotation $\gamma$, and translation ranges. Camera rotations are $ZXY$ Euler angles: $\alpha$ rotates about the patient's long axis, $\beta$ is the out-of-plane elevation, and $\gamma$ is the in-plane detector roll. Projection-mode values are sampling probabilities, drawn once per batch of rendered views. All values are sampled uniformly from the listed ranges.}
\label{tab:methods-online-drr}

\begin{tabularx}{\textwidth}{@{}>{\raggedright\arraybackslash}p{0.10\textwidth}>{\raggedright\arraybackslash}p{0.12\textwidth}>{\raggedright\arraybackslash}p{0.09\textwidth}>{\raggedright\arraybackslash}p{0.11\textwidth}>{\raggedright\arraybackslash}p{0.10\textwidth}>{\raggedright\arraybackslash}p{0.10\textwidth}>{\raggedright\arraybackslash}X@{}}
\toprule
\multicolumn{7}{@{}l}{\textit{Shared defaults}} \\
Setting & \multicolumn{6}{@{}>{\raggedright\arraybackslash}p{0.83\textwidth}@{}}{Value} \\
\midrule
Rotation range & \multicolumn{6}{@{}>{\raggedright\arraybackslash}p{0.83\textwidth}@{}}{\(\alpha\in{}\)[$-$180$^\circ$, 180$^\circ$], \(\beta\in{}\)[$-$30$^\circ$, 30$^\circ$], \(\gamma={}\)0$^\circ$ unless specified below} \\
\lightrule
Detector & \multicolumn{6}{@{}>{\raggedright\arraybackslash}p{0.83\textwidth}@{}}{256$\times$256 pixels} \\
\midrule
\multicolumn{7}{@{}l}{\textit{Dataset-specific geometry}} \\
Dataset & Isocenter & SDD (mm) & $\Delta$ (mm/pixel) & Projection mode & $\gamma$ (deg) & Translation (mm) \\
\midrule
MOOSE & random label & [1150, 1250] & [1.25, 2.25] & \makecell[l]{cone 0.9 /\\ orthographic 0.1} & 0 & \makecell[l]{\(x\in{}\)[$-$100, 100],\\ \(y\in{}\)[500, 900],\\ \(z\in{}\)[$-$100, 100]} \\
\lightrule
HaN-Seg & volume center & [1150, 1250] & [1.0, 2.0] & cone only & [$-$45, 45] & \makecell[l]{\(x\in{}\)[$-$50, 50],\\ \(y\in{}\)[500, 800],\\ \(z\in{}\)[$-$100, 100]} \\
\lightrule
Shoulder-CT & volume center & [950, 1050] & [1.9, 2.0] & cone only & [$-$45, 45] & \makecell[l]{\(x\in{}\)[$-$50, 50],\\ \(y\in{}\)[400, 600],\\ \(z\in{}\)[$-$100, 100]} \\
\lightrule
RSNAFrac & volume center & [950, 1050] & [1.0, 2.0] & cone only & [$-$45, 45] & \makecell[l]{\(x\in{}\)[$-$50, 50],\\ \(y\in{}\)[500, 800],\\ \(z\in{}\)[$-$100, 100]} \\
\lightrule
PedsCT & volume center & [1050, 1150] & [1.25, 2.0] & \makecell[l]{cone 0.9 /\\ orthographic 0.1} & 0 & \makecell[l]{\(x\in{}\)[$-$50, 50],\\ \(y\in{}\)[700, 800],\\ \(z\in{}\)[$-$50, 50]} \\
\lightrule
ElbowCT & volume center & [1050, 1150] & [1.25, 2.0] & cone only & [$-$45, 45] & \makecell[l]{\(x={}\)0,\\ \(y\in{}\)[500, 800],\\ \(z\in{}\)[$-$50, 50]} \\
\bottomrule
\end{tabularx}
\end{table*}

\subsection{DRRs generated online from CT scans}
\label{sec:appendix-dataengine-CT}

\paragraph{Randomly sampling DRRs and their labels.}
\label{sec:drr-rendering}
We render DRRs by integrating CT attenuation along source-to-detector rays~\cite{siddon1985fast,unberath2018deepdrr}, following established conventions~\cite{gopalakrishnan2026rapid}. For each view, we randomly sample the source-to-detector distance \texttt{SDD}, detector spacing \(\Delta\), C-arm rotational parameters \(\theta=[\alpha,\beta,\gamma]\), C-arm translational parameters \(\tau=[x,y,z]\), and projection geometry (cone-beam or orthographic) from dataset-specific ranges (\cref{tab:methods-online-drr}). Because \(y\) and \texttt{SDD} are drawn independently, we sort each pair so that \(y\leq\mathrm{SDD}\), and choose the per-dataset \texttt{SDD} ranges to prevent the detector plane from intersecting with the body. For a CT volume, we place the view isocenter \(c\) either at the volume's isocenter or at the centroid of a randomly sampled foreground label. As CT volumes in the MOOSE dataset span a larger FOV, we place the isocenter at the center of a randomly sampled foreground structure (\cref{tab:methods-online-drr}).

\begin{figure}[t!]
\centering
    \includegraphics[width=\columnwidth]{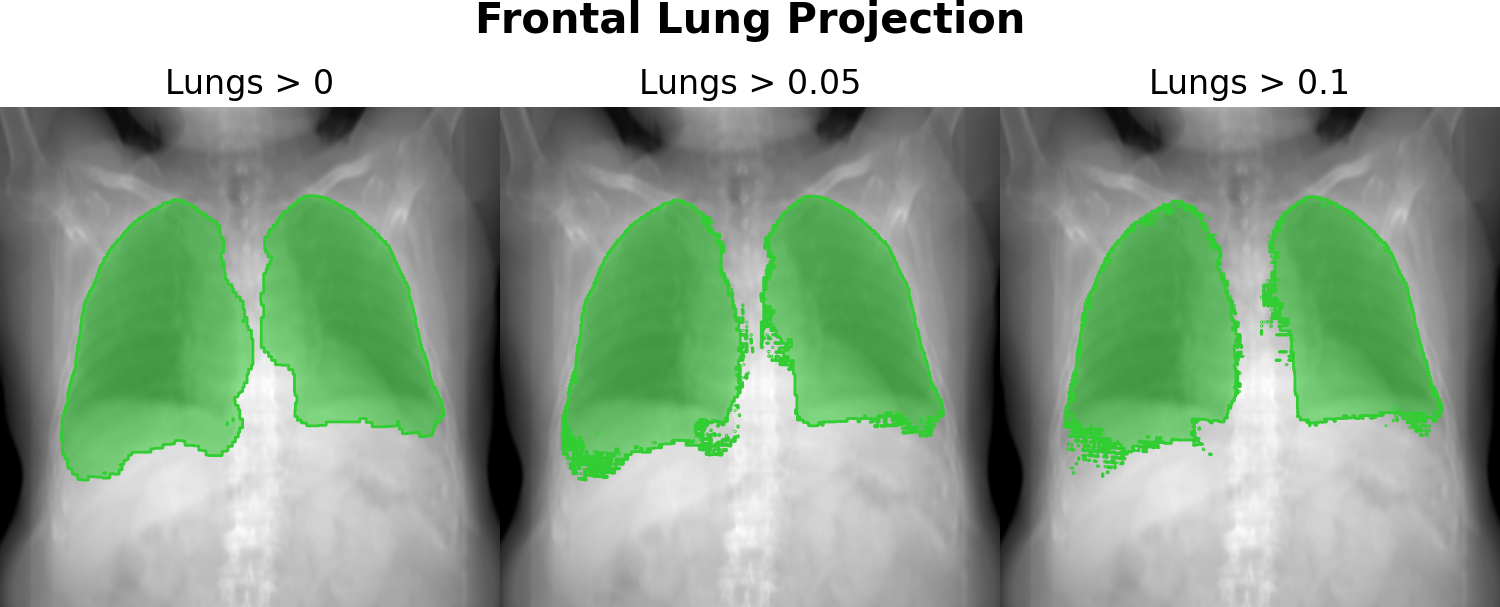}
\caption{\textbf{Lung label projection threshold.} At a threshold of 0, every pixel whose ray intersects any lung tissue is marked, including retro-cardiac and sub-diaphragmatic lung. Raising the threshold to 0.1 restricts the mask to the radiographically visible lung fields.}
\label{fig:lung-drr-seg-thresh}
\end{figure}

Segmentation masks are analytically rendered in a similar fashion to DRRs. Each projected label accumulates the attenuation contributed by casting rays passing through voxels containing that label, so thresholding the projection above 0 produces a binary 2D segmentation of that structure. The one exception to thresholding at 0 is lungs, where we use a higher threshold of 0.1 to restrict the lung mask to regions that are more aligned with existing X-ray lung masks (\cref{fig:lung-drr-seg-thresh}).

\paragraph{Label-based attenuation randomization.}
\label{sec:appendix-density-mods}

Similar to recent work in chest X-ray segmentation using CT labels~\cite{dong2025anycxr}, we use the existing segmentation supervision to simulate differences in attenuation between organs. We scale the attenuation of every voxel in each 3D foreground label \(\ell\in\{1,\ldots,L\}\) by a multiplier drawn independently per label from a truncated log-normal distribution:

\begin{equation}
\label{eq:attenuation-multiplier}
    m_\ell \sim \operatorname{TruncLogNormal}
    \left(\eta,\sigma_m;\,[0.1,\,4.0]\right),
\end{equation}
\noindent where $\eta={}$1.0 is the mode of the underlying log-normal (i.e., $\mu=\ln\eta+\sigma_m^2$) and $\sigma_m={}$1.0 is its log-space standard deviation; draws outside [0.1, 4.0] are rejected and redrawn. Attenuation randomization is applied with probability 0.5 per CT batch. We sample the per-label multipliers once per batch and shared by them amongst all rendered views.

\begin{figure*}[t!]
    \centering
    \includegraphics[width=\textwidth]{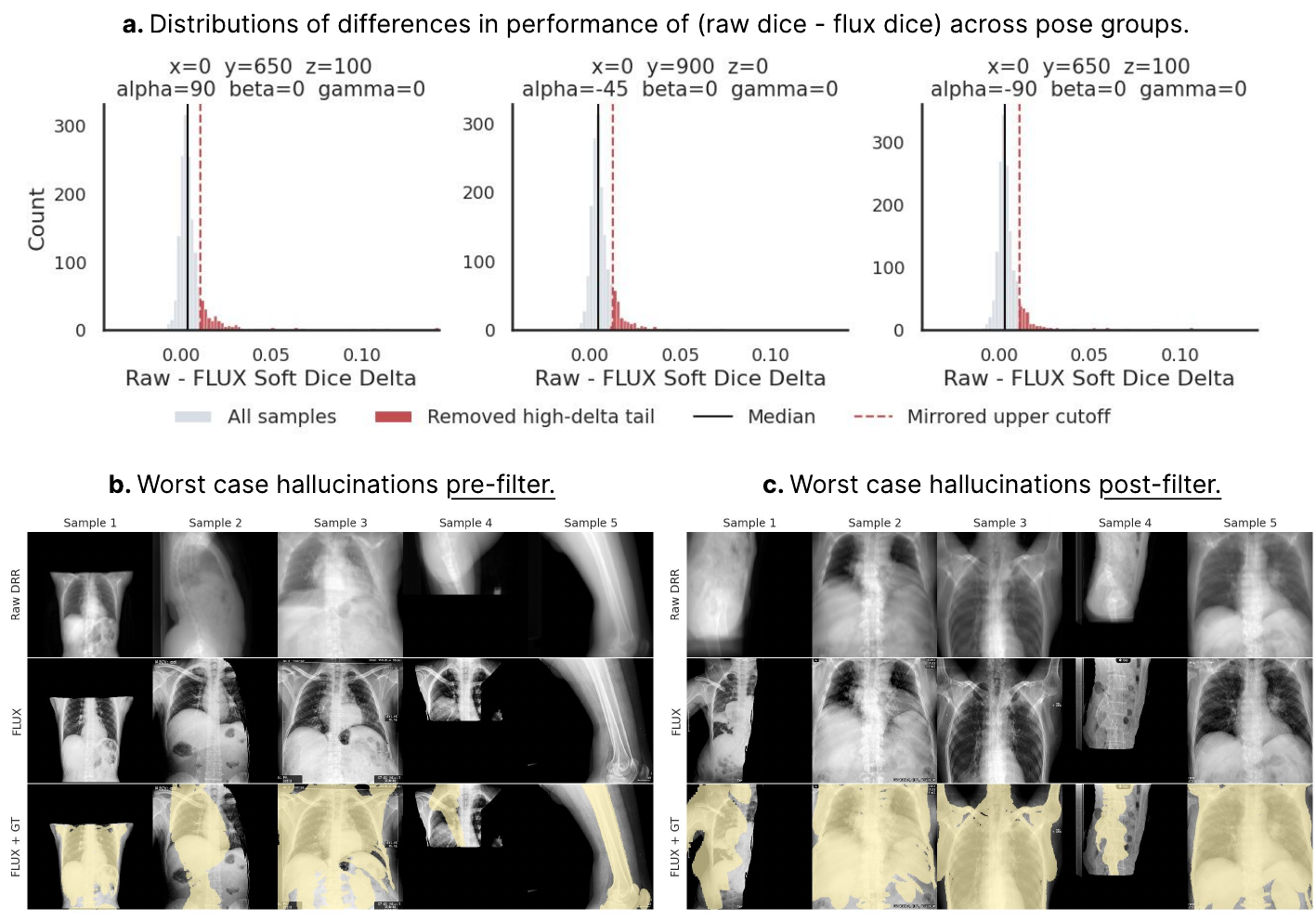}
    \caption{\textbf{Filtering \textit{Flux}-induced hallucinations.} \textbf{a,} Distributions of the per-sample SoftDice difference \(\Delta\) for three example pose groups: all samples (gray), the removed high-\(\Delta\) tail (red), the group median (solid line), and the mirrored upper cutoff \(\ell_p\) (dashed line). \textbf{b,} Worst-case hallucinations before filtering: on non-standard views, \textit{Flux} occasionally replaces the rendered anatomy with a frontal-chest-like image, so the inherited labels (yellow overlay) no longer match the image. \textbf{c,} Worst-case surviving samples after filtering, with structural hallucination substantially reduced.
    }
    \label{fig:appendix-flux-filtering}
\end{figure*}

\subsection{Diffusion-enhanced synthetic X-rays}
\label{sec:appendix-dataengine-enhanced-drrs}

DRRs can differ substantially in appearance from clinical X-rays~\cite{gao2022synthex}. To reduce this domain shift, we further generate a dataset of enhanced DRRs by applying a pretrained text-conditioned image editing diffusion model. We apply \texttt{flux.2-klein-9b}~\cite{flux-2-2025} (\textit{Flux}) to 
DRRs sampled from the whole-body MOOSE CT dataset, while retaining the DRRs' projected segmentation labels. This generative image editing is done offline so as to not bottleneck segmentation network training. This data source is constructed in three stages: rendering, enhancement, and filtering.

\paragraph{Fixed-pose rendering.}
We render 90 DRRs from each MOOSE CT volume from a fixed set of C-arm poses, producing image-segmentation pairs with high heterogeneity in anatomical FOV. Following the camera parameterization of \cref{tab:methods-online-drr}, the grid is the Cartesian product of translations \(y\in{}\)\{650, 900, 1150\}~mm and \(z\in{}\)\{$-$300, $-$200, $-$100, 0, 100, 200\}~mm with rotations \(\alpha\in{}\)\{0, $\pm$45, $\pm$90\}$^\circ$, holding \(x={}\)0~mm and \(\beta=\gamma={}\)0$^\circ$, for 3$\times$6$\times$5 $=$ 90 poses per volume. The raw render stream uses an \texttt{SDD} of 1250, detector spacing 0.6$\times$0.6, output size 1024$\times$1024, the volume center as the isocenter, and fixed (default) attenuation. During training, each enhanced DRR draw selects a subject and then one of its poses uniformly at random.

\paragraph{Diffusion model-based enhancement.}
We use \textit{Flux} to edit each DRR with 4 denoising steps, \texttt{bfloat16}, guidance scale 1.0, and the following anatomy-preserving prompt:

\begin{quote}
\itshape ``Make this synthetic X-ray (generated by a projection through a CT) look like a real X-ray. The generated image should preserve ALL anatomical structure, no shape changes at all. Add a small amount of clinical text to the image typical of X-rays but do not change the position or location of the anatomy at all. Also, add some realistic X-ray textures to the image. The output must be well registered with the input.''
\end{quote}

\noindent We found that the final sentence in the prompt was crucial for preserving alignment between the images and their labels.

\paragraph{Hallucination filtering.} 
Despite explicit instructions in the prompt, \textit{Flux} can occasionally alter projected anatomy (particularly for non-standard views), introducing disagreement between the enhanced image and inherited segmentation labels. We apply the following strategies to filter out these failures:

\subpara{Foreground filtering} 
For an enhanced DRR and its corresponding raw DRR, we set to zero enhanced pixels whose corresponding raw DRR intensity is $<$0.05. This suppresses anatomy synthesized outside of the projected body.

\subpara{Paired difference filtering}
A U-Net $\phi$, trained exclusively on raw MOOSE-derived DRRs (no real X-rays, enhanced DRRs, or other datasets), is applied to each pre-rendered DRR $x_{drr}$ and enhanced DRR $x_{flux}$, both sharing the same projected ground truth $y$, yielding predictions $\phi(x_{drr}) = \hat{y}_{drr}$ and $\phi(x_{flux}) = \hat{y}_{flux}$. We compute $s_{drr} = \text{SoftDice}(y, \hat{y}_{drr})$ and $s_{flux} = \text{SoftDice}(y, \hat{y}_{flux})$ (empty labels ignored and background included) and consider the paired difference $\Delta = s_{drr} - s_{flux}$.

We frame filtering as a per-sample, one-sided outlier rule. Within the group of enhanced DRRs that share the same pose $p$, benign enhancement draws $\Delta$ from a null distribution assumed symmetric about its median. We set an upper-tail cutoff by mirroring the empirical 2nd percentile about the median, a robust nonparametric fence in the spirit of median-based outlier rejection~\cite{leys2013detecting,rousseeuw2011robust}. A sample is kept if \(\Delta \le \ell_p\), where
\begin{equation}
\label{eq:flux-filter-cutoff}
    \ell_p = 2\,\mathrm{median}(\{\Delta\}_{p}) - q_{0.02}(\{\Delta\}_{p}),
\end{equation}
with $\mathrm{median}(\{\Delta\}_{p})$ the median difference within the group and $q_{0.02}(\{\Delta\}_{p})$ its 2nd percentile. The cutoff targets the upper 2\% of a symmetric reference distribution (\cref{fig:appendix-flux-filtering}a).

\subsection{Annotated real X-rays}
\label{sec:appendix-annotated-real-xrays}
The real X-ray datasets supplement CT-derived supervision with peripheral bone anatomy that is poorly represented in available CT datasets: HandBones, which covers diverse hand segmentations; FootBones, which has annotations for toes and metatarsals; and the MURA forearm and humerus subsets, which we annotated ourselves (\cref{sec:appendix-implementation-details}). Adding these real X-rays significantly improved performance on held-out wrist and elbow X-ray datasets (\cref{sec:appendix-ablation-training-datasets}).

\subsection{Label harmonization}
\label{sec:appendix-label-harmonization}
\method{} segments 60 structures: the skull, sternum, sacrum, and hip bones; the scapulae, clavicles, humeri, radii, ulnae, carpals, metacarpals, and hand phalanges; the femurs, patellae, tibiae, fibulae, tarsals, metatarsals, and toes; each rib pair individually (ribs 1--12); each vertebra individually (C1 -- C7, T1 -- T12, L1 -- L5); and the lungs, heart, liver, kidneys, and spleen. Bilateral structures share one channel (e.g., the left and right femur are both ``femurs'').

For each dataset, we define a mapping from its native labels to this label protocol. The mapping merges distinctions finer than ours (e.g., left and right rib labels into single per-rib channels, heart substructures into heart) and sends structures our protocol leaves out to background, dropping them from supervision.

\subsection{Data augmentation}
\label{sec:appendix-image-mask-augmentation}

To improve robustness, we apply a diverse set of augmentations during training (Table~\ref{tab:methods-augmentations}), and introduce several X-ray-specific transforms that improve generalization to real images (\cref{sec:appendix-ablation-augmentation}). \textit{Label Zoom} randomly selects one present foreground structure, computes its centroid, samples a zoom factor $z$, crops a window of size $(H/z, W/z)$ centered on that centroid, and resizes the crop back to standard image size. \textit{Letter Drop} stamps a rasterized `\texttt{L}' or `\texttt{R}' (chosen with equal probability) at a uniformly random position, with a height sampled uniformly between 3\% and 10\% of the image height and a brightness sampled from [0.8, 1.0]. \textit{Aspect Crop} chooses to either perform a horizontal or vertical crop, samples a center-crop of the alternate dimension $l$ so that the image is either $(H,l)$ or $(l,W)$, and zero-pads to maintain the image's original size $(H,W)$.

\subsection{Preprocessing}
\label{sec:appendix-preprocessing}
For CT datasets, we divided each volume into overlapping 512$\times$512$\times$256 crops to reduce disk-to-RAM transfer costs during data-loading for online DRR rendering. For each crop, we padded with HU intensities corresponding to air if the crop $z<{}$256 and limited crop-overlap to 50\%. Each crop stores the CT, label map, affine matrix, and foreground centroids used to sample randomized DRR views during training (\cref{tab:methods-online-drr}). For MOOSE, HaN-Seg, and Shoulder-CT, we ran the MOOSE-Z automated CT segmentation pipeline~\cite{sundar2022fully} to segment the volumes. For RSNAFrac, we ran TotalSegmentator~\cite{wasserthal2023totalsegmentator} to add a skull label. For PedsCT, we took the raw RTSTRUCT contours and converted them into voxel masks. Finally, for ElbowCT, we took the original label STL surface meshes and transformed them into the CT coordinate frame, sampled and voxelized, and filled them into solid masks.

For X-rays, we square-pad and resize all image-mask pairs to 256$\times$256 pixels using area interpolation for images and nearest-neighbor interpolation for masks. We split CT-derived data in fixed 90/10 train/validation splits with no test set. X-ray datasets use fixed 70/15/15 train/validation/test splits (held-out X-ray datasets use train for nnUnet training only), with three exceptions listed in \cref{tab:methods-eval-data}: RAM-W600 keeps its released partition, DeepFluoro is split by specimen (two of its six specimens per partition), and AASCE, on which no model is trained, is divided into validation and test partitions only. For all images (X-ray or DRR), we clip each to its per-image 0.5th and 99.5th intensity percentiles, and then min--max rescale each image to [0, 1]. We keep all splits subject-disjoint and fixed across experiments.

\begin{table}[t]
\centering
\scriptsize
\caption{\textbf{Training augmentation presets.} Augmentations are grouped by whether they update both image and mask or image only. The \textit{Stream} column indicates whether an operation is applied to rendered DRRs, real X-rays, or both. Rotations and shear are in degrees, kernel size, sigma, and crop sizes in pixels, translate is a fraction of image size, and noise std and intensity factors are on the [0, 1] intensity scale. Elastic alpha and sigma are fixed per-axis magnitudes rather than sampled ranges. CLAHE and gamma are mutually exclusive per sample, with the listed marginal probabilities.}
\label{tab:methods-augmentations}
\begin{tabularx}{\columnwidth}{@{}p{0.22\columnwidth}p{0.18\columnwidth}>{\raggedright\arraybackslash}X p{0.12\columnwidth}p{0.08\columnwidth}@{}}
\toprule
Operation & Parameters & Values & Stream & Prob. \\
\midrule
\multicolumn{5}{@{}l}{\textit{Image and mask}} \\
\addlinespace[0.2em]
Horizontal flip & --- & --- & both & 0.5 \\
\lightrule
Label zoom & zoom range & [1.1, 4.0] & X-ray & 0.5 \\
\lightrule
Affine & \makecell[l]{rotation\\translate\\scale\\shear} & \makecell[l]{[$-$90, 90]\\{}[$-$0.25, 0.25]\\{}[0.8, 1.2]\\{}[$-$16, 16]} & both & 0.25 \\
\lightrule
\makecell[l]{Elastic\\deformation} & \makecell[l]{kernel size\\alpha \((x,y)\)\\sigma} & \makecell[l]{127\\(12, 20)\\64} & both & 0.1 \\
\lightrule
Aspect crop & \makecell[l]{height\\width} & \makecell[l]{[64, 224]\\{}[64, 224]} & both & 0.1 \\
\midrule
\multicolumn{5}{@{}l}{\textit{Image only}} \\
\addlinespace[0.2em]
Letter drop & \makecell[l]{height\\brightness} & \makecell[l]{[0.03, 0.10]$\times H$\\{}[0.8, 1.0]} & both & 0.25 \\
\lightrule
Invert & --- & --- & both & 0.5 \\
\lightrule
CLAHE & \makecell[l]{clip limit\\grid} & \makecell[l]{[1.0, 2.0]\\8$\times$8} & both & 0.1 \\
\lightrule
Gamma & \makecell[l]{gamma\\gain} & \makecell[l]{[0.9, 1.1]\\{}[0.9, 1.1]} & both & 0.25 \\
\lightrule
Contrast & contrast & [0.7, 1.3] & both & 0.25 \\
\lightrule
\makecell[l]{Plasma\\brightness} & \makecell[l]{roughness\\intensity} & \makecell[l]{[0.3, 0.6]\\{}[0.05, 0.2]} & both & 0.1 \\
\lightrule
Sharpness & sharpness & [0.7, 1.3] & both & 0.5 \\
\lightrule
Gaussian noise & std & 0.01 & both & 0.25 \\
\bottomrule
\end{tabularx}
\end{table}

\section{Training}
\label{sec:training}

\subsection{Network architecture}
\label{sec:appendix-network-architecture}

\method{} is an ensemble of five networks (members) that share one architecture and differ only in their training-data mixture. Each member is a seven-level 2D convolutional U-Net~\cite{ronneberger2015u}, with its encoder and decoder using the same feature widths at matched resolutions, 64, 128, 256, 512, 512, 720, and 1024. Each encoder block is composed of three 3$\times$3 convolutions, each followed by a LeakyReLU activation and instance normalization. A residual connection is used around the convolutional layer, using an identity mapping when the input and output widths match, and a 1$\times$1 convolution with normalization to match feature widths otherwise. A final 1$\times$1 convolution maps the decoder features to per-channel logits with per-channel sigmoid activations used to map the logits to independent, overlapping probability maps. We also considered using a pretrained segmentation foundation model for fine-tuning with our data recipe, but did not observe a clear accuracy advantage (\cref{sec:appendix-ablation-sam2}).

MOOSE-derived data fills a fixed 75\% of the sampling budget in every member; the supplemental CT and real X-ray sources fill the remaining 25\% at the weights of \cref{tab:methods-training-data}. The members differ only in how this 75\% is divided between DRRs rendered online during training and the offline enhanced DRRs: the offline enhanced DRRs take 0, 1/3, 1/2, 2/3, or all of the MOOSE budget, with online rendering taking the rest. We found that combining online and offline rendering yields higher mean Dice than using either alone (\cref{sec:appendix-ablation-proportion-mixing}).

\subsection{Optimization}
\label{sec:appendix-training-objective}

\method{} trains for 2,000 epochs with 500 mixed-dataset loader iterations per epoch and a batch size of 16 images (either DRRs or X-rays).  AdamW~\cite{loshchilov2017decoupled} is used for optimization with an initial learning rate of 3$\times$10$^{-4}$, weight decay 0.01 on all parameters, EMA~\cite{tarvainen2017mean} with $\alpha ={}$0.9999, and a cosine learning rate scheduler stepped per epoch over the 2,000 epochs with $\eta_{min} ={}$0.0, without warm-up or gradient clipping. Inference uses the EMA weights at the final epoch of training. Importantly, we don't perform any checkpoint selection based on the results of the held-out X-ray datasets.

\paragraph{Crop sampling weights.}
To produce a DRR batch, we first sample from a CT dataset's crops with replacement under tempered inverse label-frequency weights~\cite{gupta2019lvis, mahajan2018exploring}. Each foreground structure \(c\) contributes \((N/n_c)^{\tau}\), where \(N\) is the source's crop count, \(n_c\) is the number of crops containing \(c\), and \(\tau={}\)0.5 tempers the distribution. A crop's weight is the maximum of its structures' contributions. CT crops without foreground labels, under the \method{} training label-protocol, are assigned 0 weight.

\paragraph{Training objective.}
Our training objective is an equally weighted sum of the Soft-Dice loss~\cite{milletari2016v} and binary cross-entropy (BCE). For \method{} $f$ and input image $x$, \(f(x) = z_{bci}\), target \(y_{bci}\), and probability \(p_{bci}=\sigma(z_{bci})\), the Soft-Dice term uses \(\epsilon={}\)10$^{-7}$ and the squared-denominator form
\begin{equation}
\label{eq:soft-dice}
    \mathcal{L}_{\mathrm{Dice}} =
    1 -
    \frac{2\sum_i p_{bci}y_{bci} + \epsilon}
         {\sum_i p_{bci}^2 + \sum_i y_{bci}^2 + \epsilon}
\end{equation}
Soft-Dice is restricted to foreground channels present in the image, whereas BCE supervises every channel unless a source is partially labeled, as described below.

\subpara{Partial labeling loss} 
The six data sources (RSNAFrac, PedsCT, FootBones, ElbowCT, MURA forearm, and MURA humerus) annotate only a subset of the structures visible in their images. For these, BCE is computed only on output channels that contain annotated pixels. Channels with no annotation are excluded rather than supervised as background, so structures that are present but unlabeled receive no negative signal.

\subpara{Targeted negative supervision}
Excluding empty channels also discards negative supervision on structures that are known to be absent. The byproduct is that the model hallucinates extra classes in ambiguous views. For five of the six sources we reinstate BCE on channels that are anatomically impossible in the image yet easily confused with the labeled structures: FootBones supervises the hand and forearm channels (phalanges, metacarpals, carpals, ulnae, and radii) as empty; ElbowCT, MURA forearm, and MURA humerus supervise the lower-limb channels (femurs, patellae, tibiae, and fibulae) as empty; and RSNAFrac supervises the humeri channel as empty.

\section{Evaluation}
\label{sec:evaluation}

\begin{table*}[t!]
\centering
\scriptsize
\caption{\textbf{Held-out evaluation data sources.} Counts are X-rays retained after the quality-control exclusions described in the text. Images from the same patient remain in the same split. Annotated labels are pixel-level segmentation masks, except for AASCE, whose annotations are Cobb angles derived from 68 vertebral corner landmarks (T1--L5). Mask type describes the masks used for scoring after protocol post-processing (\cref{sec:appendix-postprocessing}): structures may occupy the same pixel (overlapping) or partition the image (exclusive). Visible labels are the foreground structures exempted dataset-wide from aFPR (\cref{eq:afpr}).}
\label{tab:methods-eval-data}
\begin{tabularx}{\textwidth}{@{}lp{0.16\textwidth}>{\centering\arraybackslash}p{0.03\textwidth}>{\centering\arraybackslash}p{0.03\textwidth}>{\centering\arraybackslash}p{0.03\textwidth}>{\centering\arraybackslash}p{0.10\textwidth}>{\raggedright\arraybackslash}X>{\raggedright\arraybackslash}X@{}}
\toprule
Dataset & Description & Train & Val & Test & Mask type & Annotated labels & Visible labels \\
\midrule
\multicolumn{8}{@{}l}{\textit{Primary segmentation evaluation}} \\
\addlinespace[0.2em]
DarwinCVD19~\cite{v7labs-covid19} & Chest X-ray & 4,443 & 952 & 952 & Exclusive & Lungs & Clavicles, heart, humeri, kidneys, liver, \textbf{lungs}, ribs 1--12, scapulae, skull, spleen, sternum, vertebrae C1--L5 \\
\lightrule
DeepFluoro~\cite{grupp2020automatic} & Fluoroscopic X-ray & 70 & 79 & 213 & Overlapping & Femurs, hips, lumbar spine, sacrum & \textbf{Femurs}, \textbf{hips}, lumbar spine, sacrum, vertebrae L1--L5 \\
\lightrule
ElbowLat~\cite{ionspace-elbowlat} & Lateral elbow X-ray & 419 & 91 & 91 & Overlapping & Humeri, radii, ulnae & \textbf{Humeri}, \textbf{radii}, \textbf{ulnae} \\
\lightrule
HipRay~\cite{gut2021x} & Pelvic X-ray & 97 & 21 & 21 & Exclusive & Femurs, hips & \textbf{Femurs}, \textbf{hips}, sacrum \\
\lightrule
LowerLimbs~\cite{bone-identifier-1rey5_dataset} & Pelvis and lower extremity & 39 & 9 & 8 & Overlapping & Femurs, fibulae, tibiae & \textbf{Femurs}, \textbf{fibulae}, hips, metatarsals, patellae, sacrum, tarsals, \textbf{tibiae}, toes, vertebrae L1--L5 \\
\lightrule
RAM-W600~\cite{yang2025ram} & Hand/wrist X-ray & 425 & 69 & 124 & Overlapping & Carpals, metacarpals, radii, ulnae & \textbf{Carpals}, \textbf{metacarpals}, phalanges, \textbf{radii}, \textbf{ulnae} \\
\lightrule
PedsTorso~\cite{monchbot1_thoracoabdominal_2026} & Pediatric X-ray & 54 & 12 & 12 & Exclusive & Lungs, thoracolumbar spine & Clavicles, femurs, heart, hips, humeri, kidneys, liver, \textbf{lungs}, radii, ribs 1--12, sacrum, scapulae, skull, spleen, sternum, \textbf{thoracolumbar spine}, ulnae, vertebrae C1--L5 \\
\lightrule
VinDr-Rib~\cite{nguyen2021vindr} & Chest X-ray & 171 & 37 & 37 & Overlapping & Ribs 1--10 & Clavicles, heart, humeri, kidneys, liver, lungs, \textbf{ribs 1--12}, scapulae, skull, spleen, sternum, vertebrae C1--L5 \\
\midrule
\multicolumn{8}{@{}l}{\textit{Downstream tasks}} \\
\addlinespace[0.2em]
AASCE~\cite{wang2021evaluation} & Spinal AP X-ray & 0 & 262 & 218 & --- & Cobb angles & --- \\
\lightrule
BTXRD~\cite{yao2025radiograph} & Musculoskeletal X-ray & 1,305 & 281 & 281 & Exclusive & Bone tumors & --- \\
\lightrule
MTDDH~\cite{qi2025mtddh} & Pediatric pelvic X-ray & 632 & 138 & 135 & Overlapping & Femoral head, ilium, ischium, pubis & --- \\
\bottomrule
\end{tabularx}
\end{table*}

X-ray segmentation evaluation is challenging due to the scarcity of diverse segmentation datasets available online. We made a best-effort attempt to gather a diverse selection of eight X-ray segmentation datasets that cover a broad range of target anatomy and acquisition settings (\cref{tab:methods-eval-data}). These collectively cover the trunk and both limbs, spanning lungs (DarwinCVD19), ribs (VinDr-Rib), spine (PedsTorso, DeepFluoro), pelvis and hip (HipRay, DeepFluoro), femur, tibia, and fibula (LowerLimbs), upper limbs (ElbowLat), and the carpals, metacarpals, and forearm bones of the hand and wrist (RAM-W600). They span standard and oblique radiography and intra-operative fluoroscopy (DeepFluoro), frontal and lateral acquisitions (ElbowLat), and both adult and pediatric anatomy (PedsTorso). 

In five of the eight sets, the evaluation masks retain spatial overlap, so one pixel can carry several labels (Mask type in \cref{tab:methods-eval-data}). \method{} and the generalist baselines each emit a separate mask per structure, and the dataset-specific nnU-Nets are trained in region-based mode on these five datasets (\cref{sec:appendix-baselines}), so no method is restricted to a mutually exclusive partition when scoring them.

All of the main reported results, and the ablations, are scored on the test partitions, which were held out during model development. The validation partitions were used only for development decisions, such as choosing the \method{} (single) network for the ablations and finetuning (\cref{tab:ablation-enhanced-drrs-proportion}). The pretrained \method{} members and every ablation model are evaluated at their final training epoch without checkpoint selection (\cref{sec:appendix-training-objective}). All evaluation annotations were sourced from publicly released datasets and annotation projects.

\subpara{Quality control and splits} The per-image split assignment of every source, together with each excluded image and its reason, is released with the dataset (\url{https://huggingface.co/datasets/VictorButoi/flexray-data}). For sources we cannot redistribute, the lists are keyed by the original filenames. Before assigning splits, we removed exact-duplicate images, which would otherwise place the same radiograph in several partitions, and images whose annotation contained no foreground after mapping into our label protocol. These affected DarwinCVD19 (33 duplicates, 108 empty masks, and 16 annotations whose image is absent from the public download) and ElbowLat (11 empty masks). We additionally excluded a small number of individual images whose annotations we judged unreliable on manual review: one in HipRay, five in LowerLimbs, four DeepFluoro frames with questionable ground-truth poses reported upstream, one AASCE radiograph whose landmarks fall outside the image, and three MTDDH images with a corrupt file or an out-of-bounds polygon. No other evaluation dataset had exclusions. All criteria were fixed before any model was scored and apply identically to \method{} and every baseline.

\subsection{Protocol post-processing}
\label{sec:appendix-postprocessing}

In order to compare our predictions, and those of our baselines, to the held-out X-ray ground-truth labels, we apply minimal deterministic post-processing to all predictions to align them with our evaluation sets. Each adjustment is applied identically to every method. All rules were derived from each dataset's annotation documentation and inspection of its training-split annotations, fixed before scoring the evaluation partitions, and were not tuned on their performance. 

For quantitative evaluation on lung datasets (DarwinCVD19 and PedsTorso), we set lung probabilities to zero where there is predicted liver or spleen to restrict the predictions to visible contours of the lungs. For DeepFluoro lumbar-spine and PedsTorso thoracolumbar-spine, we take the pixelwise maximum in probability space over the corresponding vertebra channels (L1--L5, and T1--T12 together with L1--L5, respectively). For PedsTorso and HipRay, we collapse both the prediction and the reference into a single mutually exclusive overlay under a fixed per-dataset priority order (thoracolumbar spine over lungs for PedsTorso, femurs over hips for HipRay).

\subsection{Metrics}
\label{sec:appendix-metrics}

\paragraph{Segmentation quality.} We evaluate segmentation overlap quality with both the Dice similarity coefficient~\cite{dice1945measures}, which measures the spatial overlap between a ground-truth mask and a predicted mask, and 95th-percentile Hausdorff distance (HD95)~\cite{huttenlocher1993comparing}, which measures boundary error in pixels. For every boundary pixel of one mask we take the Euclidean distance to the nearest boundary pixel of the other, and HD95 is the larger of the two directed 95th percentiles; boundaries are each mask minus its one-pixel erosion. A structure present in the ground truth but absent from the thresholded prediction is excluded from that image's HD95 average. Dice and HD95 are computed per structure and averaged per image. A structure whose ground truth occupies less than 0.1\% of the image area is dropped from that image, so that vanishingly small structures do not contribute to its score.

\paragraph{Anatomical false-positive rate (aFPR).}

To quantify hallucinations in \method{} predictions, we introduce the anatomical false-positive rate (aFPR): the average number of predicted foreground classes per image that are anatomically incompatible with the X-ray dataset. For an image \(x\) from dataset \(d\), let \(\mathcal{P}(x)\) be the classes whose predicted probability exceeds 0.5 anywhere in the image, \(\mathcal{G}(x)\) the classes in its ground truth, and \(\mathcal{I}_d\) the classes visible in the radiographs of dataset \(d\), annotated or not (visible labels in \cref{tab:methods-eval-data}). Then
\begin{equation}
\label{eq:afpr}
    \mathrm{aFPR}(x) = \left|\mathcal{P}(x) \setminus \left(\mathcal{G}(x) \cup \mathcal{I}_d\right)\right|.
\end{equation}

\paragraph{Mean target registration error (mTRE).}
We evaluate 2D/3D registration accuracy with the mean target registration error. Given a set of 3D fiducial landmarks \(\{\mathbf{t}_k\}_{k=1}^{K}\) defined in the CT coordinate frame (\(K={}\)14 per DeepFluoro specimen), a recovered camera pose \(\hat{p}\), and the ground-truth pose \(p^\ast\),
\begin{equation}
\label{eq:mtre}
\mathrm{mTRE}(\hat{p}, p^\ast) \;=\; \frac{1}{K}\sum_{k=1}^{K}\bigl\lVert T_{\hat{p}}(\mathbf{t}_k) - T_{p^\ast}(\mathbf{t}_k)\bigr\rVert_2,
\end{equation}
where \(T_p\) maps a point into the camera coordinate frame of pose \(p\). mTRE is reported in millimeters; lower is better.

\paragraph{Detection average precision (AP50).} For detection targets we score whether each annotated lesion is localized rather than how well it is delineated. Candidate detections are the 4-connected components of at least four pixels in which the predicted lesion probability exceeds 0.5, each represented by its enclosing box and scored by its mean lesion probability. Candidates are matched one-to-one to ground-truth boxes in decreasing order of confidence, a match requiring an intersection-over-union of at least 0.5, with at most 100 candidates per image, using the reference COCO implementation~\cite{lin2014microsoft}. AP50 is the area under the 101-point interpolated precision--recall curve. Higher is better.

\begin{table}[h!]
\centering
\scriptsize

\caption{\textbf{Test-time augmentation preset.} Each augmented view draws the operations below independently, in the order listed, with the per-draw probability in Prob. CLAHE and gamma are mutually exclusive per draw.}
\label{tab:methods-tta}
\begin{tabularx}{\columnwidth}{@{}p{0.27\columnwidth}p{0.22\columnwidth}>{\raggedright\arraybackslash}X p{0.10\columnwidth}@{}}
\toprule
Operation & Parameters & Values & Prob. \\
\midrule
Horizontal flip & --- & --- & 0.5 \\
\lightrule
Invert & --- & --- & 0.5 \\
\lightrule
CLAHE & \makecell[l]{clip limit\\grid} & \makecell[l]{[1.0, 2.0]\\8$\times$8} & 0.1 \\
\lightrule
Gamma & \makecell[l]{gamma\\gain} & \makecell[l]{[0.9, 1.1]\\{}[0.9, 1.1]} & 0.25 \\
\lightrule
Contrast & contrast & [0.7, 1.3] & 0.25 \\
\lightrule
Sharpness & sharpness & [0.7, 1.3] & 0.5 \\
\lightrule
Gaussian noise & std & 0.01 & 0.25 \\
\bottomrule
\end{tabularx}
\end{table}

\subsection{Test-time augmentation}
\label{sec:appendix-tta}

\begin{figure}[t]
\centering
    \includegraphics[width=\columnwidth]{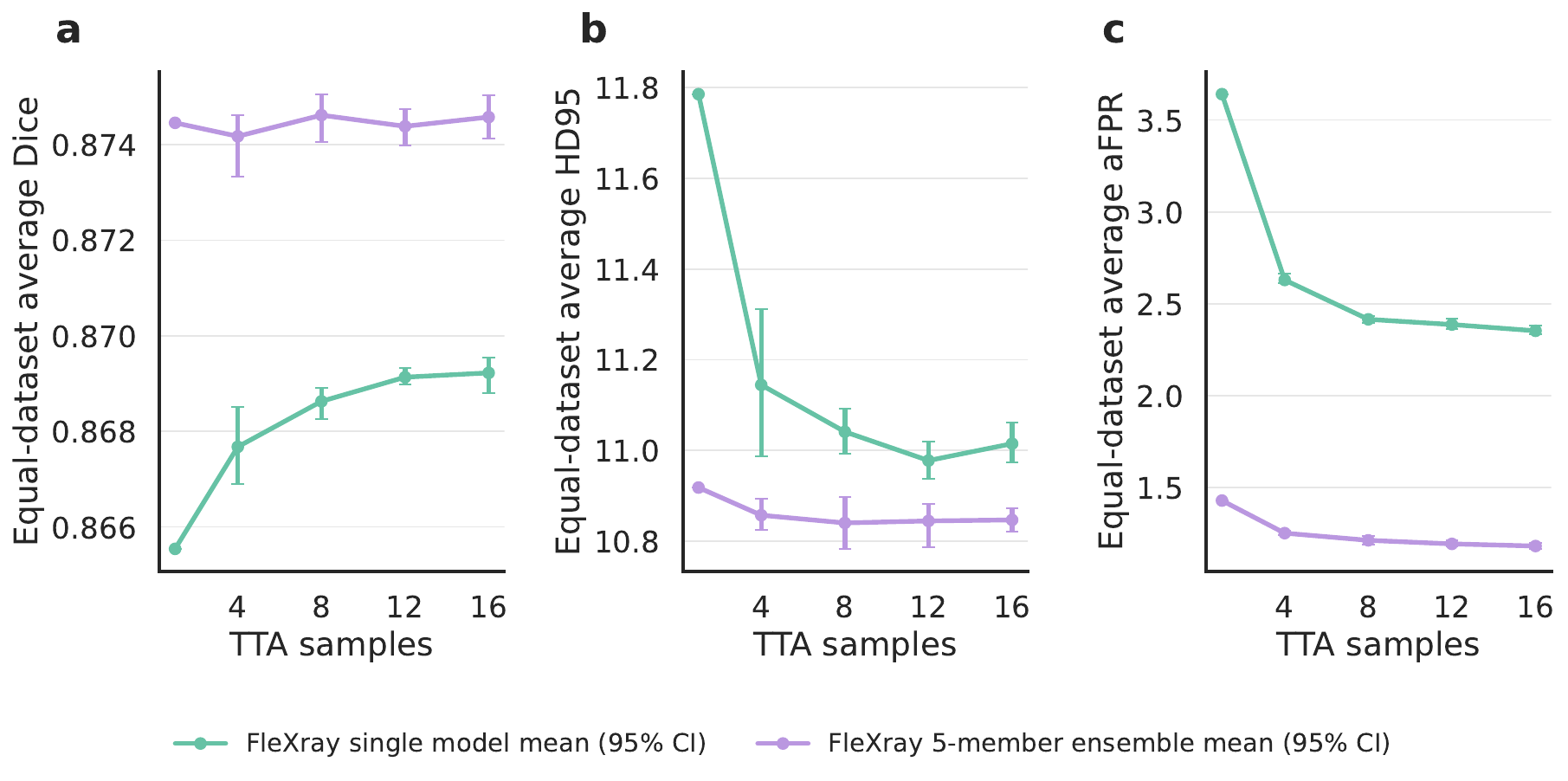}
\caption{\textbf{TTA sample sweep.} Equal-dataset mean \textbf{a,} Dice, \textbf{b,} HD95, and \textbf{c,} aFPR versus the number of TTA samples for \method (single) and \method{}, computed over the eight held-out real-radiograph segmentation datasets in \cref{tab:methods-eval-data}. Each line is the per-method mean over four inference seeds (40--43) with 95\% confidence intervals over seeds. Higher Dice and lower HD95 and aFPR are better.}
\label{fig:appendix-tta-sweep}
\end{figure}

To improve robustness and reduce false-positive predictions (\cref{fig:appendix-tta-sweep}), we wrap \method{} in a test-time augmentation (TTA) module that aggregates predictions over several stochastically perturbed views of each image~\cite{shanmugam2020and, shanmugam2021better}. The first pass uses the unaugmented image and the remaining passes are independent draws from the TTA preset of \cref{tab:methods-tta}. We ensure that every ensemble member receives the same set of $K$ augmented views:

\begin{equation}
\hat{y}_{ens} = \frac{1}{5} \sum_{i=1}^5 \left[ \frac{1}{K} \sum_{k=1}^{K} T_k^{-1} (f_i(T_k(x))) \right],
\end{equation}

\noindent where $T_k$ is the $k^{th}$ augmentation view and $T_k^{-1}$ is its necessary inverse operation (identity except for views with left/right flips). 

The number of TTA samples, \(K={}\)16, was fixed on the validation partitions of \cref{tab:methods-eval-data} before the final models were trained. \Cref{fig:appendix-tta-sweep} reports this frozen setting on the test partitions, sweeping \(K\) only to show how Dice, HD95, and aFPR vary around it. Ensembling and TTA change Dice and HD95 very little. From the \method{} (single) model without TTA to \method{}, equal-dataset mean Dice rises from 0.866 to 0.875 and HD95 falls from 11.8 to 10.8 pixels. Both substantially reduce hallucinated structures. For the \method{} (single) model, 16 TTA samples lower aFPR (\cref{eq:afpr}) from 3.64 to 2.35 classes per image. Ensembling the five members without TTA lowers it to 1.43. Combining the two yields 1.18, roughly one third of the single-pass rate.

\subsection{Baselines}
\label{sec:appendix-baselines}

We compare against general-purpose X-ray segmentation methods that, like \method{}, did not train on the held-out real X-ray evaluation sets. As a reference, we also trained a set of dataset-specific nnU-Net models on each of the X-ray evaluation segmentation datasets. We evaluated only labels represented by a method's released output vocabulary. For our baselines, we applied the TTA scheme that they considered as part of their method (no TTA if they did not use it). Table~\ref{tab:appendix-baseline-efficiency} summarizes their computational footprints, and \cref{tab:appendix-baseline-dice} reports the full per-dataset comparison (both in \cref{sec:appendix-full-baseline-comparison}).

\paragraph{PAXray~\cite{seibold2023accurate}.}
We used the authors' ResNet50-UNet pretrained checkpoint which predicts 159 output channels. Each X-ray was replicated to three channels, normalized with ImageNet statistics, and resized to 512$\times$512 to match their training.

\paragraph{TotalSegmentator2D~\cite{alshenoudy2025leveraging}.} 
We used the released ensemble of models for cardiac, muscle, organ, rib, and vertebral groups, yielding 117 native foreground channels. Images were processed with the nnU-Net plans and preprocessing stored with the released models. For DeepFluoro only, we inverted each image's intensities within its per-image dynamic range before applying the TotalSegmentator2D preprocessing. Inference uses nnU-Net's default flip-mirroring test-time augmentation.

\paragraph{FluoroSAM~\cite{killeen2025fluorosam}.} 
FluoroSAM differs from other baselines and \method{} as it is an interactive-segmentation method. We used the released Swin-L checkpoint at its 448$\times$448 training resolution. Prompt names were derived from the foreground labels in our shared protocol. For each scored non-bilateral label, we supplied its text prompt and one positive point at the center of each connected ground-truth component. For bilateral structures, we issued separate left- and right-specific text prompts with one positive ground-truth-derived point per available side and combined the two score maps by pixelwise maximum.

\paragraph{Efficiency measurement.} We time per-image model execution at batch size 1 in FP32 on an NVIDIA V100 at each method's native input resolution: 256$\times$256 for \method{} and for each of the five TotalSegmentator2D anatomy-group models, 512$\times$512 for PAXray, and 448$\times$448 for FluoroSAM. Input preparation and dataset-specific postprocessing are outside the timer. Ensemble timings include prediction averaging, and TTA timings additionally include augmentation, normalization, inverse alignment, and averaging across views. We select one test image from each of the eight primary segmentation evaluation datasets in \cref{tab:methods-eval-data}. For each image, latency is the mean of 10 CUDA-synchronized runs after 5 warmup runs, and memory is the peak allocation during a post-warmup run. We report mean latency across the eight images with an image-level bootstrap confidence interval. FluoroSAM's timing covers one image-encoder pass followed by 62 text-prompt encoder and mask-decoder passes, covering the 60 protocol structures plus lumbar spine and thoracolumbar spine, without point prompts.

\paragraph{Dataset-specific networks.} We trained independent nnU-Net~\cite{isensee2021nnu} networks for each evaluation dataset in \cref{tab:methods-eval-data}. We use nnU-Net's default configuration and 1,000-epoch schedule, without dataset-specific hyperparameter tuning or fold ensembling. Where annotated structures overlap (Mask type in \cref{tab:methods-eval-data}), we train nnU-Net in its region-based mode. Mutually exclusive datasets use the standard softmax configuration. Inference uses nnU-Net's default flip-mirroring test-time augmentation.

\section{Downstream tasks}
\label{sec:appendix-downstream}

\subsection{Cobb-angle estimation}
\label{sec:appendix-cobb}

\paragraph{Splits and development.} AASCE is partitioned by patient into a validation split (262 images; 39, 107, and 116 low, moderate, and severe curves) and a test split (218 images; 33, 100, and 85).

\paragraph{Procedure.} We apply the same procedure with identical settings to \method{} and every baseline. We threshold the prediction channels for T1--T12 and L1--L5, the 17 vertebrae annotated in AASCE, at 0.5, assigning overlapping pixels to the more superior vertebra. For each vertebra, we retain the largest connected component and discard it if its area is below 0.1\% of the image. We order the surviving vertebrae superior to inferior by mask centroid. Images with fewer than three surviving vertebrae are excluded from that method's error statistics and counted toward its failure rate.

\paragraph{Cobb-angle prediction.} Following Pan et al.~\cite{pan2019computer}, we approximate vertebral end-plate orientations from the spinal centerline connecting consecutive mask centroids. For each interior vertebra, the end-plate direction is the axial bisector of the normals to its two adjacent centerline segments, treating the normals as undirected lines. The first and last end-plate directions are horizontal. We represent these directions by unit vectors pointing toward image right and report their largest pairwise angular separation as the predicted major Cobb angle, following the challenge's reference procedure~\cite{wang2021evaluation}. The reference major angle is the maximum of the three provided AASCE angles.

\subsection{2D/3D registration}
\label{sec:appendix-registration}

We use the semantic information estimated by \method to inject additional supervision into 2D/3D registration protocols via a bidirectional Chamfer distance~\cite{tenenbaum1977parametric, borgefors1986distance}. We build upon \xvr~\cite{gopalakrishnan2026rapid}, which registers a preoperative CT to an intraoperative X-ray \(x\) by minimizing an image similarity loss $\mathcal L_{\mathrm{img}}$ between a DRR $I_p$ rendered from camera pose $p$ and $x$ with respect to $p$. We focus on DeepFluoro as a representative study.

\paragraph{Per-structure Chamfer loss.} 
Let \(\mathcal{S}\) be the set of structures in the pelvis (i.e., hips, lumbar vertebrae, and sacrum). For each structure \(s\in\mathcal{S}\), let \(M_s\subseteq\mathcal{D}\) be its \method{}-predicted mask, where \(\mathcal{D}\) denotes the set of detector pixels and \(u\in\mathcal{D}\) a pixel. 

\subpara{Forward term} We pre-compute the 2D Euclidean distance transform \(D_s(u)\) of \(M_s\), representing the distance from \(u\) to the nearest pixel of \(M_s\) in mm. Next, as the preoperative CT is labeled, the renderer partitions the DRR into per-structure channels \(I_p^{s}(u)\). For structures with nonzero projected mass, the forward term integrates the unit-normalized projected mass of \(s\) over all detector pixels against \(D_s\):
\begin{equation}
\ell^{\mathrm{fwd}}_s(p) = \frac{\sum_{u\in\mathcal{D}} I_p^{s}(u)\,D_s(u)}{\sum_{u\in\mathcal{D}} I_p^{s}(u)}.
\end{equation}

\subpara{Reverse term} We pre-compute the 3D Euclidean distance field \(\Phi_s(\cdot)\) from every voxel to \(s\) in mm. For every pixel in the predicted mask $u \in M_s$, we sample points $\mathbf v_u \in \mathbb R^{K\times3}$ along the camera ray and reduce:
\begin{equation}
\ell^{\mathrm{rev}}_s(p) = \frac{1}{|M_s|}\sum_{u\in M_s}\bigl\langle \alpha_u,\,\Phi_s(\mathbf{v}_u)\bigr\rangle,
\end{equation}
where \(\alpha_u=\operatorname{softmax}\bigl(-\Phi_s(\mathbf{v}_u)/\tau\bigr)\) with temperature \(\tau={}\)\SI{2}{mm} and  \(K={}\)300. The reverse term penalizes poses for which the \method{} mask's rays are far from the 3D structure. 

\subpara{Why distance transforms} Both terms owe their robustness to a property of Euclidean distance transforms. The distance fields $D_s$ and $\Phi_s$ have unit-magnitude spatial gradients almost everywhere outside their respective structures, regardless of distance, satisfying the Eikonal equation, $\lVert\nabla \Phi_s\rVert = 1$~\cite{delfour2011shapes}. The reverse term uses these gradients through its softmax-weighted aggregation, and the forward term weights displaced projected mass linearly in its distance to the predicted mask. An image similarity objective, by contrast, carries no signal once the DRR and X-ray no longer overlap, so the Chamfer terms provide informative gradients even for very poor initial pose estimates.

\paragraph{Optimization.} Each term is averaged over the structures \(\mathcal{S}^{+}\subseteq\mathcal{S}\) that have a non-empty predicted segmentation mask, with weights \(\omega_s\) proportional to the area of \(M_s\) at the coarsest resolution and normalized to sum to one, so that large, well-observed structures dominate:
\begin{equation}
\begin{aligned}
\mathcal{L}_{\mathrm{fwd}}(p) &= \sum_{s\in\mathcal{S}^{+}}\omega_s\,\ell^{\mathrm{fwd}}_s(p),\\
\mathcal{L}_{\mathrm{rev}}(p) &= \sum_{s\in\mathcal{S}^{+}}\omega_s\,\ell^{\mathrm{rev}}_s(p).
\end{aligned}
\end{equation}
The final registration objective is
\begin{equation}
\label{eq:registration-objective}
\mathcal{L}(x, p) = \mathcal L_{\mathrm{img}}(x, I_p) + w\,\bigl(\mathcal{L}_{\mathrm{fwd}}(p) + \mathcal{L}_{\mathrm{rev}}(p)\bigr).
\end{equation}

\noindent Gradient-based optimization follows the multistage protocol of \xvr~\cite{gopalakrishnan2026rapid}, with the Chamfer weights $w$ following a per-stage schedule tuned for each initialization. From the coarse initialization, where image similarity is uninformative, the weight is \(w={}\)1 at the two coarse stages and \(w={}\)0.05 at the finest, so the coarse stages pull the pose into the mask-consistent basin while the final fit is determined by image similarity. From the foundation-model initialization, we use a weak schedule of \(w={}\)0.05, 0.05, and 0.01 to provide moderate supervision while safeguarding against divergence. 

\paragraph{\method{} predicted masks.} We apply the default \method{} configuration to each frame after the same cropping and log-linearization used by the registration, and threshold predictions at 0.5. Segmentation masks are predicted by \method{} once per frame and are shared across every initialization and method.

\paragraph{Evaluation protocol.} DeepFluoro~\cite{grupp2020automatic} provides ground-truth camera poses and 3D fiducial landmarks for six cadaver specimens. We evaluate on the 213 frames of the two held-out test specimens (\cref{tab:methods-eval-data}). Each frame is registered from two initializations: a coarse, clinically practical pose constructed from anatomical landmarks (Method 1 in~\citet{grupp2020automatic}), and the pose predicted by the whole-body foundation pose-regression model of \xvr~\cite{gopalakrishnan2026rapid}, which is not trained on DeepFluoro. Because a single radiograph does not disambiguate the anterior--posterior orientation, the foundation initialization is run in both orientations and the result with the lower final objective (\cref{eq:registration-objective}) is kept for each method, so the two methods can start from different orientations of the same frame and their initial errors differ slightly (76.6 versus 76.1~mm median). 

From each initialization, we run \xvr with and without the Chamfer terms and score the recovered pose by the mean target registration error over the specimen's \(K={}\)14 anatomical landmarks (bilateral pelvic landmarks defined in the CT frame by Grupp et al.~\cite{grupp2020automatic}; mTRE, \cref{sec:appendix-metrics}), calling a registration successful below \SI{10}{mm}. Image-similarity registration takes 3--5 seconds per frame. The Chamfer terms raise this to about 9 seconds from the foundation initialization and 28 seconds from the coarse initialization, whose strong weights keep the optimizer from stopping early.

\subsection{Data-efficient finetuning}
\label{sec:appendix-finetune}

All fine-tuned networks are initialized from the final checkpoint of \method{} (single). We replace the final 1$\times$1 output convolution with a randomly initialized head sized to the target protocol, and fine-tune end-to-end. 

\paragraph{Optimization.} Each run trains for 250 epochs of 50 iterations at a batch size of 16 and 256$\times$256 resolution. We use AdamW with a peak learning rate of 3$\times$10$^{-4}$, weight decay 3$\times$10$^{-5}$, a 5-epoch linear warm-up from 0.1$\times$ the peak, and cosine decay thereafter, in 32-bit precision and without EMA. We use the same objective as \method{} pre-training (\cref{sec:appendix-training-objective}). For every run we keep the checkpoint with the highest mean foreground Dice on the validation split and report on the test split. To ensure a fair comparison, during training we apply the default 2D augmentation pipeline of nnU-Net v2~\cite{isensee2021nnu} (rotation, scaling, Gaussian noise and blur, brightness, contrast, simulated low resolution, and gamma), restricting rotation to $\pm$15$^\circ$ and using the same left--right flip as TTA.

\paragraph{Training subsets.} For few-shot configurations, we draw subsets from the training split of each dataset (\cref{tab:methods-eval-data}). Within a repeat, every smaller subset is a prefix of the larger ones. Budgets below 100\% use three repeats drawn with different seeds. \method{} and the from-scratch nnU-Net comparators (\cref{sec:appendix-baselines}) read the identical subset manifests, validate on the same validation split, and are scored on the same test split.

\paragraph{MTDDH.} The head has five channels: background, ilium, pubis, ischium, and femoral head. The femoral head overlaps the acetabular bones, so MTDDH is an overlapping dataset (\cref{tab:methods-eval-data}) and its nnU-Net comparator is trained in region-based mode on the same four foreground labels. The dataset's native femur annotation is mapped to background for both \method{} and nnU-Net because \method{} already segments the femur. Subsets are selected at the patient level by a seeded random permutation, at 1\%, 5\%, 10\%, 50\%, and 100\% of the training split (6, 32, 63, 316, and 632 patients).

\paragraph{BTXRD.} The head has two channels, background and tumor, and all images carry a tumor. Subsets cover 1\%, 5\%, 10\%, 50\%, and 100\% of the training split (13, 65, 130--131, 653, and 1,305 images). Because BTXRD spans three centers and nine diagnosis categories with skewed frequencies, subsets are stratified rather than drawn uniformly. Cases are added in an order that preserves the full split's center and diagnosis proportions at every budget.

BTXRD is scored by detection AP50 (\cref{sec:appendix-metrics}), the endpoint reported by the dataset's authors~\cite{yao2025radiograph}. Ground truth is the expert's bounding rectangle for each of the 354 tumors in the test split, transformed with the same padding and resizing as the image. Both \method{} and nnU-Net are scored on softmax probabilities averaged over the image and its horizontal reflection. AP50 is computed per network and averaged over training repeats. Tumor-size quartiles are fixed once on the training split by box area as a fraction of the padded image; following the COCO convention, unmatched candidates outside a quartile's size range are ignored rather than counted as false positives. Because every BTXRD radiograph contains a tumor, precision is measured on tumor-positive images only and does not estimate specificity on normal radiographs.

\addtocontents{toc}{\protect\setcounter{tocdepth}{-1}}%
\crefalias{section}{appendix}
\crefalias{subsection}{appendix}
\crefalias{subsubsection}{appendix}

\section{Additional results}
\label{sec:appendix-additional-results}

\begin{table*}[t]
\centering
\scriptsize
\setlength{\tabcolsep}{2pt}
\renewcommand{\arraystretch}{1.25}
\newcommand{\diceci}[3]{\makecell{#1\\{\tiny[#2,\,#3]}}}
\newcommand{\bestdiceci}[3]{\makecell{\textbf{#1}\\{\tiny[#2,\,#3]}}}
\caption{\textbf{Per-dataset Dice comparison against baselines.} Per-example mean label Dice, shown as mean [95\% CI] (\cref{sec:appendix-statistics}), on the eight held-out real X-ray test sets (\(n\) images each). \method{} (ours) is the released five-member ensemble with 16-sample test-time augmentation (\cref{sec:appendix-network-architecture,tab:methods-tta}); the other \method{} rows remove TTA, the ensemble, or both. Dashes mark datasets outside a method's label coverage. Bold marks \method{} (ours) and baselines not significantly different from it. Paired comparisons include only \method{} (ours) versus each supported generalist baseline, with Holm correction within each dataset. nnU-Net is greyed as an in-domain reference trained on each dataset's own annotations. Mean weights the eight datasets equally and is shown only for methods covering all of them.}
\label{tab:appendix-baseline-dice}
\begin{tabular}{@{}l c c c c c c c c c@{}}
\toprule
Method & \makecell{RAM-W600\\(\textit{n}~$=$~124)} & \makecell{LowerLimbs\\(\textit{n}~$=$~8)} & \makecell{ElbowLat\\(\textit{n}~$=$~91)} & \makecell{HipRay\\(\textit{n}~$=$~21)} & \makecell{DeepFluoro\\(\textit{n}~$=$~213)} & \makecell{PedsTorso\\(\textit{n}~$=$~12)} & \makecell{VinDr-Rib\\(\textit{n}~$=$~37)} & \makecell{DarwinCVD19\\(\textit{n}~$=$~952)} & \makecell{Mean\\Dice} \\
\midrule
PAXray                                      & --- & --- & --- & --- & --- & \diceci{0.570}{0.499}{0.645} & \bestdiceci{0.751}{0.733}{0.769} & \diceci{0.888}{0.884}{0.892} & --- \\
\lightrule
TotalSeg2D                                  & --- & --- & --- & \diceci{0.478}{0.423}{0.531} & \diceci{0.371}{0.357}{0.386} & \diceci{0.719}{0.603}{0.808} & \diceci{0.695}{0.672}{0.715} & \diceci{0.906}{0.903}{0.908} & --- \\
\lightrule
FluoroSAM                                   & \diceci{0.239}{0.230}{0.249} & \diceci{0.416}{0.300}{0.520} & \diceci{0.182}{0.161}{0.202} & \diceci{0.313}{0.277}{0.353} & \diceci{0.711}{0.701}{0.722} & \diceci{0.308}{0.220}{0.385} & \diceci{0.091}{0.080}{0.103} & \diceci{0.527}{0.519}{0.534} & 0.348 \\
\midrule
\method{} (single)                          & \diceci{0.945}{0.944}{0.947} & \diceci{0.831}{0.717}{0.920} & \diceci{0.854}{0.836}{0.871} & \diceci{0.937}{0.934}{0.941} & \diceci{0.904}{0.898}{0.908} & \diceci{0.811}{0.780}{0.839} & \diceci{0.724}{0.705}{0.742} & \diceci{0.918}{0.915}{0.921} & 0.866 \\
\lightrule
\method{} (single) + 16 TTA               & \diceci{0.946}{0.944}{0.947} & \diceci{0.835}{0.723}{0.920} & \diceci{0.864}{0.847}{0.879} & \diceci{0.936}{0.932}{0.940} & \diceci{0.910}{0.905}{0.914} & \diceci{0.817}{0.789}{0.843} & \diceci{0.726}{0.704}{0.745} & \diceci{0.922}{0.920}{0.925} & 0.870 \\
\lightrule
\method{} (no TTA)                          & \diceci{0.948}{0.947}{0.950} & \diceci{0.836}{0.714}{0.929} & \diceci{0.871}{0.855}{0.883} & \diceci{0.938}{0.934}{0.942} & \diceci{0.913}{0.909}{0.917} & \diceci{0.830}{0.805}{0.853} & \diceci{0.736}{0.718}{0.753} & \diceci{0.923}{0.920}{0.926} & 0.874 \\
\lightrule
\textbf{\method{} (ours)}                   & \bestdiceci{0.949}{0.947}{0.950} & \bestdiceci{0.834}{0.715}{0.931} & \bestdiceci{0.871}{0.855}{0.884} & \bestdiceci{0.938}{0.934}{0.942} & \bestdiceci{0.915}{0.912}{0.919} & \bestdiceci{0.831}{0.806}{0.855} & \bestdiceci{0.737}{0.716}{0.756} & \bestdiceci{0.924}{0.921}{0.927} & \textbf{0.875} \\
\midrule
\textcolor{gray}{nnU-Net (in-domain)}       & \textcolor{gray}{\diceci{0.988}{0.987}{0.988}} & \textcolor{gray}{\diceci{0.928}{0.885}{0.959}} & \textcolor{gray}{\diceci{0.921}{0.907}{0.931}} & \textcolor{gray}{\diceci{0.982}{0.979}{0.985}} & \textcolor{gray}{\diceci{0.866}{0.856}{0.875}} & \textcolor{gray}{\diceci{0.864}{0.823}{0.898}} & \textcolor{gray}{\diceci{0.800}{0.772}{0.825}} & \textcolor{gray}{\diceci{0.961}{0.959}{0.962}} & \textcolor{gray}{0.914} \\
\bottomrule
\end{tabular}
\end{table*}

\subsection{Full quantitative quality comparison against baselines}
\label{sec:appendix-full-baseline-comparison}

\Cref{tab:appendix-baseline-dice} gives the per-dataset results behind \cref{fig:quant-eval}. On the wrist, elbow, and lower-limb datasets, FluoroSAM is the only released generalist with sufficient label coverage, and \method{} exceeds it by 0.42--0.71 Dice even though FluoroSAM receives a ground-truth-derived text prompt and click for every structure (\cref{sec:appendix-baselines}). On the hip and on fluoroscopy, TotalSegmentator2D covers the labels but reaches only 0.478 and 0.371, the latter after per-image intensity inversion (0.026 if using the native polarity for DeepFluoro). The margins are smallest on adult chest radiographs, the domain both chest generalists target. On VinDr-Rib, PAXray is numerically higher but not significantly different from \method{} (0.751 vs.\ 0.737; \textit{n}$=$37 images; \textit{p}$=$0.146). On PedsTorso, \method{} significantly exceeds TotalSegmentator2D (0.831 vs.\ 0.719; \textit{n}$=$12; \textit{p}$=$0.031). On DarwinCVD19, \method{} leads TotalSegmentator2D by 0.018 Dice (0.924 vs.\ 0.906; \textit{n}$=$952; \textit{p}$<$0.001).

The in-domain nnU-Nets exceed \method{} on seven datasets by 0.03--0.09 Dice. The exception is DeepFluoro, where the nnU-Net trained on the 70 frames of two subjects reaches 0.866 on the two held-out subjects, against 0.915 for \method{} (\textit{n}$=$213; \textit{p}$<$0.001).

The four \method{} configurations separate the contributions of ensembling and test-time augmentation. Averaging the five members raises the equal-dataset mean Dice from 0.866 to 0.874; 16-sample TTA adds a further 0.004 to \method{} (single) and 0.001 to the ensemble. These gains come at the cost of additional computation (\cref{tab:appendix-baseline-efficiency}). \method{} (single) is of the same order as PAXray in size and latency (101.4M vs.\ 73.4M parameters; 14.7 vs.\ 15.5~ms per image), TotalSegmentator2D accumulates 167.5M parameters and 23.8~ms across its five anatomy-group models, and FluoroSAM (268.7M parameters) needs 623.7~ms for 62 text prompts, a cost that grows with the number of structures queried. The released ensemble uses five times as many parameters (507.0M) and takes 74.2~ms per image; TTA increases its latency to 1.24~s, while peaking at 2.36~GB of GPU memory. \method{} (single) without TTA therefore retains 99\% of the released configuration's accuracy at approximately 1/84 of its latency.

\begin{table}[h]
\centering
\scriptsize
\setlength{\tabcolsep}{3.2pt}
\renewcommand{\arraystretch}{1.2}
\caption{\textbf{Computational efficiency of \method{} and the generalist baselines.} Parameter count, peak GPU memory, and per-image latency of model execution at batch size 1 in FP32 on an NVIDIA V100 at each method's native input resolution (\cref{sec:appendix-baselines}). Latency is the mean over eight test images, one per primary segmentation evaluation dataset, with a 95\% confidence interval (\cref{sec:appendix-statistics}). TotalSeg2D sums its five sequential anatomy-group models; FluoroSAM uses 62 text prompts without point prompts, and its latency grows with the number prompted. Ensemble and TTA timings include prediction averaging, with TTA timings also including augmentation and inverse alignment. \method{} rows are as in \cref{tab:appendix-baseline-dice}.}
\label{tab:appendix-baseline-efficiency}

\begin{tabularx}{\columnwidth}{@{}X c c c@{}}
\toprule
Method & \makecell{Params\\(M)} & \makecell{Peak GPU mem.\\(GB)} & \makecell{Latency (ms)\\{\tiny mean [95\% CI]}} \\
\midrule
PAXray & 73.4 & 0.80 & \makecell{15.5\\{\tiny[15.4,\,15.8]}} \\
\lightrule
TotalSeg2D (5 subnetworks) & 167.5 & 0.89 & \makecell{23.8\\{\tiny[23.5,\,24.1]}} \\
\lightrule
FluoroSAM & 268.7 & 1.64 & \makecell{623.7\\{\tiny[615.1,\,632.3]}} \\
\midrule
\method{} (single) & 101.4 & 0.62 & \makecell{14.7\\{\tiny[14.7,\,14.8]}} \\
\lightrule
\method{} (single) + 16 TTA & 101.4 & 0.68 & \makecell{288.4\\{\tiny[287.5,\,289.4]}} \\
\lightrule
\method{} (no TTA) & 507.0 & 2.31 & \makecell{74.2\\{\tiny[74.1,\,74.3]}} \\
\lightrule
\textbf{\method{} (ours)} & 507.0 & 2.36 & \makecell{1242.3\\{\tiny[1241.3,\,1243.1]}} \\
\bottomrule
\end{tabularx}
\end{table}

\subsection{Ablations}
\label{sec:ablations}

We trained several different versions of the \method{} (single) network to evaluate different parts of our training pipeline. The ablations reuse the \method{} training recipe (\cref{sec:training}) except for the components under study and the SAM2-specific input resolutions and optimization settings described in \cref{sec:appendix-ablation-sam2}. All ablation models are evaluated on the test partitions of the held-out real-X-ray suite (\cref{tab:methods-eval-data}) with the same 16-sample test-time augmentation.

\subsubsection{Mixing online and offline rendering performs best}
\label{sec:appendix-ablation-proportion-mixing}

\method{} draws its MOOSE-derived supervision from two streams: DRRs rendered online during training (\cref{sec:appendix-dataengine-CT}) and the offline enhanced DRRs dataset (\cref{sec:appendix-dataengine-enhanced-drrs}). We trained five networks that hold the combined budget at 75\% (all other streams at the weights of \cref{tab:methods-training-data}) while sweeping the online/offline split: 75/0, 50/25, 37.5/37.5, 25/50, and 0/75.

The three mixed splits performed about the same (equal-dataset mean Dice 0.870--0.874; \cref{tab:ablation-enhanced-drrs-proportion}), and both extremes were worse. Online rendering alone (75/0) reached 0.858 and dropped to 0.705 on LowerLimbs: online DRRs are rendered from 512$\times$512$\times$256 CT crops (\cref{sec:appendix-preprocessing}) and therefore cannot simulate the far, whole-limb views in that dataset. Offline rendering alone (0/75) reached an equal-dataset mean Dice of 0.852 and was lowest on four of the eight datasets, consistent with overfitting to the fixed offline renders. Because of this, we chose the balanced 37.5/37.5 split as the representative \method{} (single) from its validation-partition score.

\begin{table*}[t]
\centering
\scriptsize
\setlength{\tabcolsep}{3pt}
\renewcommand{\arraystretch}{1.25}
\newcommand{\dicecell}[3]{\makecell{#1\\{\tiny[#2,\,#3]}}}
\caption{\textbf{Online rendering fraction sweep.} MOOSE-derived data accounts for 75\% of total training steps (\cref{tab:methods-training-data}). The first column gives the fraction of this budget assigned to online DRRs, with the remaining fraction assigned to offline enhanced DRRs. We report per-dataset mean Dice for the five networks and their ensemble (the released \method{}) on the test partitions of the eight held-out real X-ray datasets (\(n\) is the number of test images). Each cell reports the per-example mean Dice with its 95\% confidence interval (\cref{sec:appendix-statistics}). Mean Dice averages the per-dataset means with equal weight. All rows use 16-sample TTA.}
\label{tab:ablation-enhanced-drrs-proportion}
\begin{tabular}{@{}l c c c c c c c c c@{}}
\toprule
\makecell{Online rendering\\fraction} & \makecell{RAM-W600\\(\textit{n}~$=$~124)} & \makecell{LowerLimbs\\(\textit{n}~$=$~8)} & \makecell{ElbowLat\\(\textit{n}~$=$~91)} & \makecell{HipRay\\(\textit{n}~$=$~21)} & \makecell{DeepFluoro\\(\textit{n}~$=$~213)} & \makecell{PedsTorso\\(\textit{n}~$=$~12)} & \makecell{VinDr-Rib\\(\textit{n}~$=$~37)} & \makecell{DarwinCVD19\\(\textit{n}~$=$~952)} & \makecell{Mean\\Dice} \\
\midrule
1 & \dicecell{0.947}{0.946}{0.948} & \dicecell{0.705}{0.398}{0.924} & \dicecell{0.875}{0.860}{0.887} & \dicecell{0.946}{0.942}{0.950} & \dicecell{0.911}{0.907}{0.915} & \dicecell{0.824}{0.795}{0.852} & \dicecell{0.724}{0.699}{0.747} & \dicecell{0.928}{0.924}{0.931} & 0.858 \\
\lightrule
2/3 & \dicecell{0.947}{0.945}{0.948} & \dicecell{0.852}{0.755}{0.923} & \dicecell{0.871}{0.855}{0.883} & \dicecell{0.936}{0.933}{0.940} & \dicecell{0.915}{0.912}{0.919} & \dicecell{0.826}{0.803}{0.848} & \dicecell{0.727}{0.703}{0.747} & \dicecell{0.919}{0.916}{0.922} & 0.874 \\
\lightrule
1/2 & \dicecell{0.946}{0.944}{0.947} & \dicecell{0.835}{0.723}{0.920} & \dicecell{0.864}{0.847}{0.879} & \dicecell{0.936}{0.932}{0.940} & \dicecell{0.910}{0.905}{0.914} & \dicecell{0.817}{0.789}{0.843} & \dicecell{0.726}{0.704}{0.745} & \dicecell{0.922}{0.920}{0.925} & 0.870 \\
\lightrule
1/3 & \dicecell{0.946}{0.944}{0.948} & \dicecell{0.846}{0.742}{0.924} & \dicecell{0.865}{0.848}{0.878} & \dicecell{0.934}{0.929}{0.938} & \dicecell{0.910}{0.906}{0.914} & \dicecell{0.832}{0.812}{0.853} & \dicecell{0.710}{0.689}{0.730} & \dicecell{0.920}{0.917}{0.923} & 0.870 \\
\lightrule
0 & \dicecell{0.947}{0.945}{0.949} & \dicecell{0.765}{0.619}{0.892} & \dicecell{0.834}{0.806}{0.857} & \dicecell{0.927}{0.924}{0.931} & \dicecell{0.893}{0.889}{0.898} & \dicecell{0.818}{0.790}{0.844} & \dicecell{0.722}{0.700}{0.740} & \dicecell{0.913}{0.910}{0.916} & 0.852 \\
\midrule
Ensemble & \dicecell{0.949}{0.947}{0.950} & \dicecell{0.834}{0.715}{0.931} & \dicecell{0.871}{0.855}{0.884} & \dicecell{0.938}{0.934}{0.942} & \dicecell{0.915}{0.912}{0.919} & \dicecell{0.831}{0.806}{0.855} & \dicecell{0.737}{0.716}{0.756} & \dicecell{0.924}{0.921}{0.927} & 0.875 \\
\bottomrule
\end{tabular}
\end{table*}

\subsubsection{Finetuned SAM2 encoders show no significant advantage over a U-Net}
\label{sec:appendix-ablation-sam2}

To test whether the visual prior of a natural-image foundation model improves transfer to X-rays, we fully finetune SAM2~\cite{ravi2024sam} image encoders in place of the \method{} U-Net. We consider all four SAM2 Hiera variants, Tiny, Small, Base+, and Large (26.9M, 34.0M, 68.7M, and 212.2M parameters), each paired with a linear segmentation head over the 60-channel label protocol. SAM2 and \method{} (single) share the complete data engine, label harmonization, augmentation pipeline, training objective, and evaluation pipeline, including the same 16-sample test-time augmentation.

The SAM2 backbones are finetuned end-to-end at a reduced learning rate (3$\times$10$^{-5}$, versus 3$\times$10$^{-4}$ for the segmentation head). SAM2 uses AdamW weight decay of 0.05 for both the backbone and head, compared with 0.01 for \method{} (single). Tiny, Small, and Base+ use 896$\times$896 inputs and 2,000 epochs; Hiera Large uses 1024$\times$1024 inputs and 1,000 epochs, compared with 256$\times$256 inputs and 2,000 epochs for \method{} (single). In preliminary sweeps we trained every Hiera scale at backbone learning rates of 3$\times$10$^{-4}$ (uniform with the segmentation head), 3$\times$10$^{-5}$, and 10$^{-5}$, holding the head learning rate fixed. The 10$\times$ reduced backbone rate of 3$\times$10$^{-5}$ performed best, with a ranking that was stable across encoder scales, so all final runs finetune the full, unfrozen encoder at this rate.

\begin{figure*}[t]
\centering
    \includegraphics[width=\textwidth]{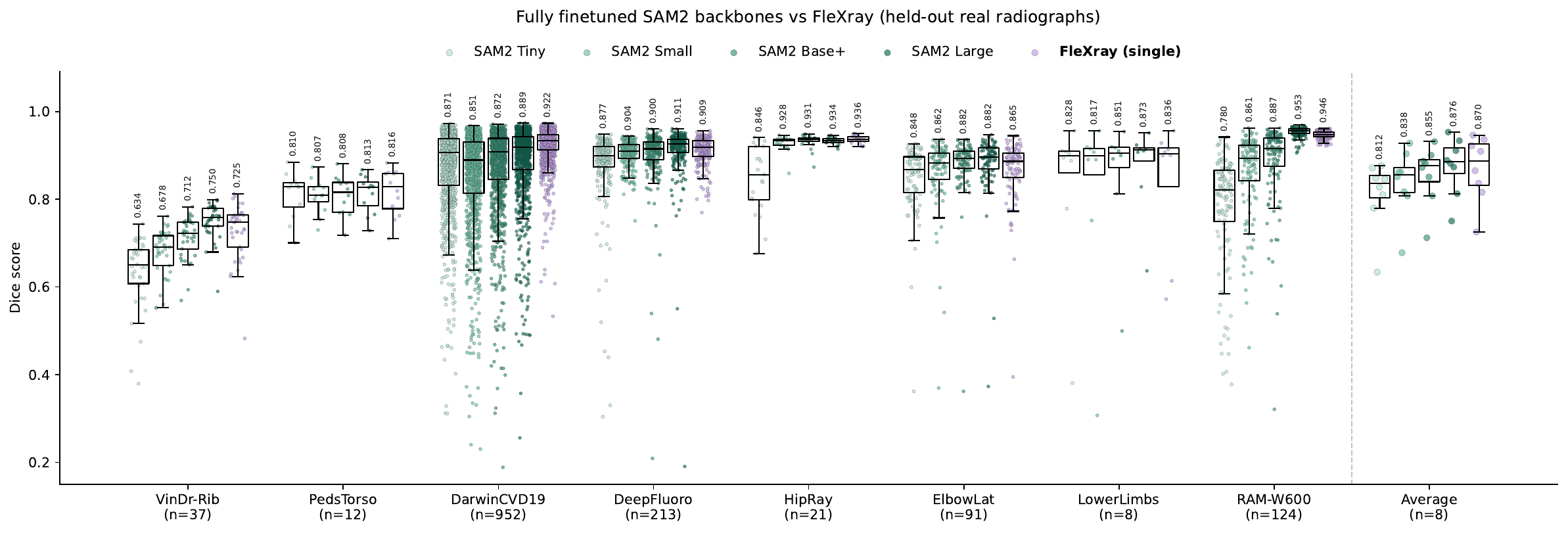}
\caption{\textbf{Per-dataset Dice of finetuned SAM2 encoders and \method{} (single).} Per-image Dice on the test partitions of the eight held-out real-radiograph datasets for fully finetuned SAM2 Hiera encoders with a linear segmentation head and \method{} (single). Points are images; boxes show median and quartiles, whiskers 1.5$\times$ the inter-quartile range; numbers give per-dataset means. Right, one point per dataset (\textit{n}$=$8); numbers give the equal-dataset mean Dice.}
\label{fig:ablation-sam2}
\end{figure*}

Accuracy rose monotonically with encoder scale. The equal-dataset mean Dice was 0.812 [0.753, 0.857] for Hiera Tiny, 0.838 [0.781, 0.885] for Small, 0.855 [0.805, 0.896] for Base+, and 0.876 [0.829, 0.916] for Large, against 0.870 [0.817, 0.915] for \method{} (single) (\textit{n}$=$1,458 test images across the eight held-out datasets; \cref{fig:ablation-sam2}). On the same images, Tiny and Small trailed \method{} (single) by 0.058 [0.023, 0.098] and 0.031 [0.008, 0.056] Dice, whereas Base+ trailed it by 0.014 [$-$0.005, 0.034] and Large exceeded it by 0.006 [$-$0.009, 0.023], intervals that both include zero (paired hierarchical bootstrap of the equal-dataset mean difference). Given our data engine, we find no statistically detectable accuracy advantage for the finetuned SAM2 encoders over a from-scratch U-Net in these experiments (\cref{sec:appendix-network-architecture}).

\subsubsection{Image-space augmentation drives transfer to real X-rays}
\label{sec:appendix-ablation-augmentation}

Transfer from rendered to real X-rays depends on appearance robustness that the model can only acquire through augmentation. Our data-engine injects variation at two stages: at render time, through per-label attenuation randomization, and in image space, through the augmentation preset of \cref{tab:methods-augmentations}. To attribute the transfer gain between these stages, we trained four regimes that add components cumulatively: (i) a base regime with neither; (ii) attenuation randomization alone; (iii) attenuation plus intensity inversion, which we isolate because radiographs are displayed in both polarities, making inversion the single largest appearance shift in the preset; and (iv) attenuation plus the complete preset, which is \method (single). All other training choices are unchanged.

\begin{figure}[h!]
\centering
    \includegraphics[width=\columnwidth]{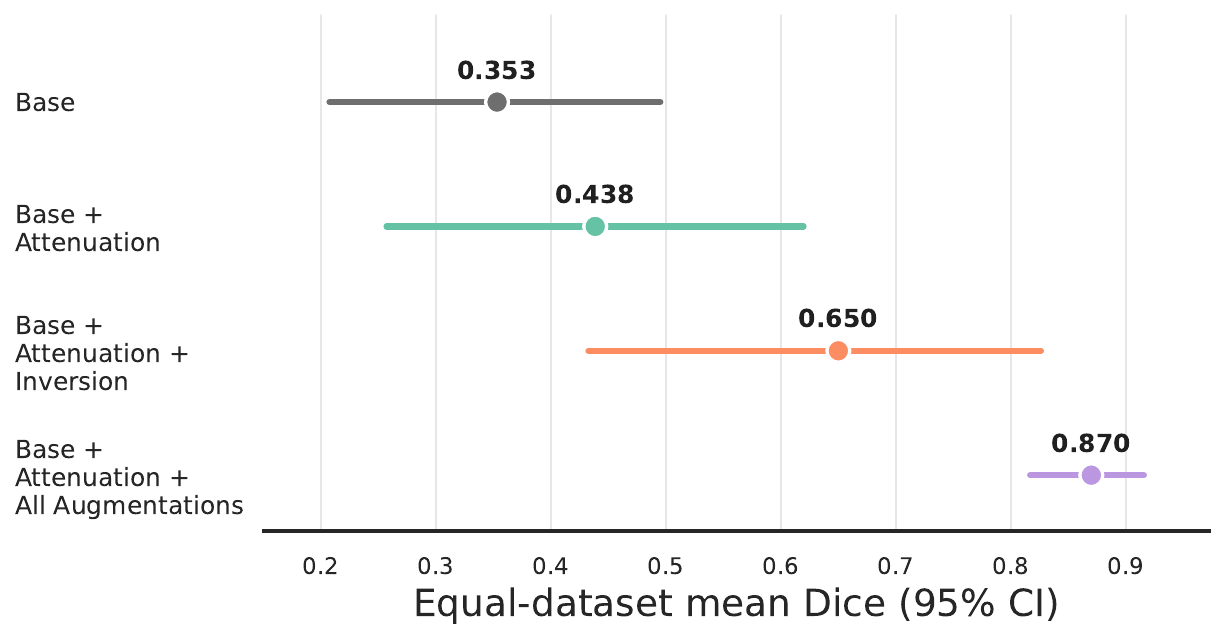}
\caption{\textbf{Cumulative augmentation ablation.} Equal-dataset mean Dice on the test partitions of the eight held-out datasets (\textit{n}$=$1,458 images) as attenuation randomization, intensity inversion, and the complete augmentation preset are progressively added. Base trains with neither attenuation randomization nor image-space augmentation. Error bars are hierarchical 95\% confidence intervals over datasets and then images.}
\label{fig:ablation-augmentation}
\end{figure}

The equal-dataset mean Dice rose from 0.353 for the base regime to 0.438 with attenuation randomization alone, to 0.650 with intensity inversion added, and to 0.870 with the complete preset (\textit{n}$=$1,458 test images across the eight held-out datasets; \cref{fig:ablation-augmentation}). Per-dataset paired comparisons localize the three additions; every difference named below is significant. Attenuation randomization alone improved only the adult chest radiographs and the pelvic fluoroscopy (+0.15 Dice on DarwinCVD19 and +0.07 on DeepFluoro) and was indistinguishable from the base regime on the other six datasets. Adding inversion improved seven of the eight datasets (all but the eight-image LowerLimbs), most on the hip and pelvic views (+0.46 on HipRay and +0.39 on DeepFluoro). Only the complete preset recovered the wrist and elbow datasets, which stayed below 0.3 Dice under every partial regime (+0.90 on RAM-W600 and +0.57 on ElbowLat over the inversion regime), whereas on the rib and pediatric datasets the inversion regime already matched it. Attenuation randomization therefore contributes modestly in isolation, whereas the image-space augmentations account for most of the transfer, and no single image-space operation suffices: inversion recovers only part of the full preset's gain, and the remainder is concentrated on the peripheral anatomy. We retain the complete preset in the released model.

\begin{table}[t]
\centering
\scriptsize
\setlength{\tabcolsep}{6pt}
\renewcommand{\arraystretch}{1.25}
\newcommand{\dicecell}[3]{\makecell{#1\\{\tiny[#2,\,#3]}}}
\newcommand{\bestdicecell}[3]{\makecell{\textbf{#1}\\{\tiny[#2,\,#3]}}}
\caption{\textbf{Training-source ablation.} Per-example mean label Dice, shown as mean [95\% CI] (\cref{sec:appendix-statistics}), for the four training-source mixtures on the test sets of RAM-W600 and ElbowLat (\(n\) images each), the two datasets showing significant source effects of at least 0.01 Dice (\cref{sec:appendix-ablation-training-datasets}). Source effects compare the complete mixture with each version omitting a source family; these three mixtures divide their MOOSE-derived budget equally between online and enhanced offline DRRs. Mean Dice weights all eight held-out datasets equally. Bold marks the highest mean and mixtures not significantly different from it on that dataset.}
\label{tab:ablation-training-datasets}
\begin{tabular}{@{}l c c c@{}}
\toprule
Mixture & \makecell{RAM-W600\\(\textit{n}~$=$~124)} & \makecell{ElbowLat\\(\textit{n}~$=$~91)} & \makecell{Mean\\Dice} \\
\midrule
Online MOOSE only & \dicecell{0.004}{0.002}{0.006} & \dicecell{0.435}{0.384}{0.485} & 0.697 \\
\lightrule
Without real X-rays & \dicecell{0.004}{0.003}{0.005} & \dicecell{0.541}{0.483}{0.597} & 0.712 \\
\lightrule
Without supplemental CT & \bestdicecell{0.948}{0.946}{0.949} & \dicecell{0.842}{0.820}{0.860} & 0.868 \\
\lightrule
Complete mixture (flagship) & \dicecell{0.946}{0.944}{0.948} & \bestdicecell{0.865}{0.849}{0.879} & \textbf{0.870} \\
\bottomrule
\end{tabular}
\end{table}

\subsubsection{Real X-rays provide training coverage for peripheral radiographs}
\label{sec:appendix-ablation-training-datasets}

\method{} draws supervision from three families of sources: the MOOSE-derived streams (online and offline renders), five supplemental CT datasets, and four manually annotated real X-ray sources (\cref{tab:methods-training-data}). We train three mixtures that divide their MOOSE-derived sampling budget equally between online rendering and offline enhanced DRRs: (i) the complete mixture (75\%/21\%/4\% MOOSE-derived/supplemental CT/real X-ray); (ii) the mixture without real X-rays (79\%/21\%/0\%); and (iii) the mixture without supplemental CT (96\%/0\%/4\%). When a source family is removed, its sampling share is returned equally to the two MOOSE-derived streams, keeping the number of optimization steps fixed. We quantify each source family's contribution by comparing the complete mixture with the corresponding omission. An additional reference is trained on online MOOSE renders alone.

We focus on source effects that are significant (\cref{sec:appendix-statistics}) and at least 0.01 Dice. Under this criterion, the real X-ray sources (535 annotated training radiographs, 4\% of the sampling budget) improve two datasets: RAM-W600, whose wrists are otherwise essentially unsegmented (0.004 to 0.946, \textit{n}$=$124), and ElbowLat (+0.325 Dice, \textit{n}$=$91; both \textit{p}$<$0.001; \cref{tab:ablation-training-datasets}). Including real X-rays raises the equal-dataset mean Dice from 0.712 to 0.870. The supplemental CT datasets (503 annotated training subjects, 21\% of the budget) provide an additional gain on ElbowLat (+0.024 Dice, \textit{n}$=$91, \textit{p}$<$0.001). No other dataset meets both criteria in either comparison. These comparisons show the importance of real X-ray annotations for the wrist and elbow datasets, with supplemental CT providing a smaller additional benefit on the elbow. The held-out suite does not evaluate the skull and shoulder anatomy targeted by some supplemental CT sources, so this ablation does not measure their contribution to those regions.

\begin{figure*}[t!]
\centering
    \includegraphics[width=\textwidth]{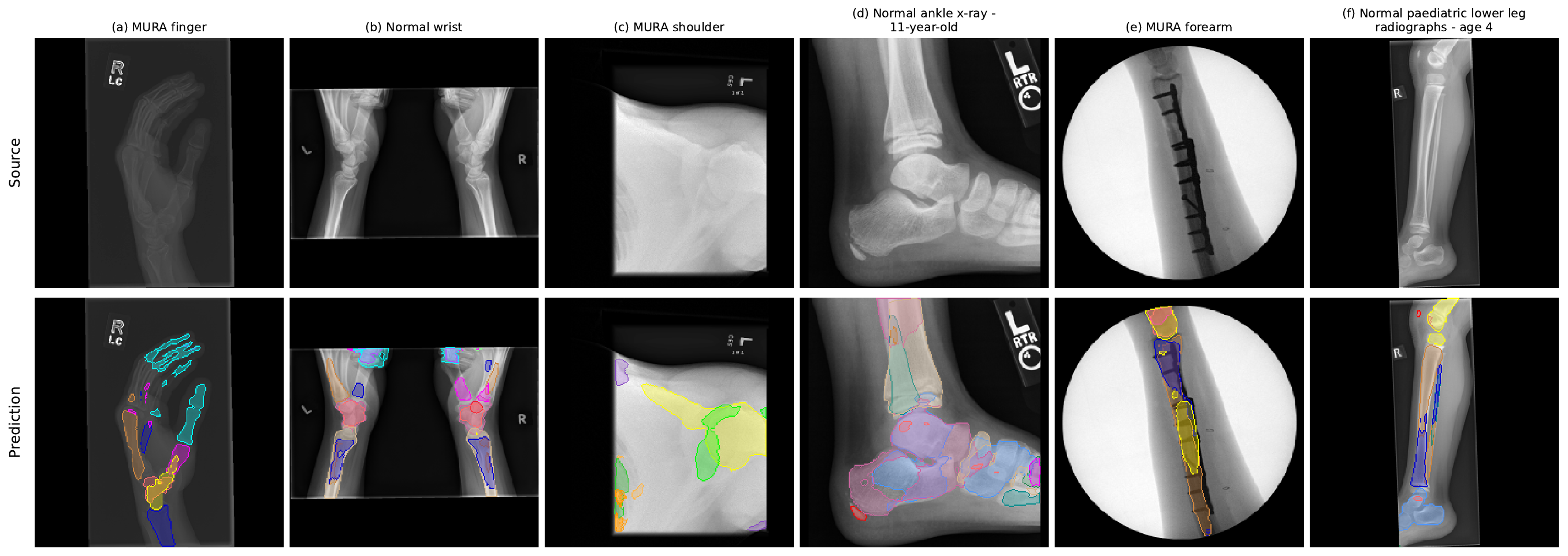}
\caption{\textbf{Illustration of failure cases.} Predictions of the released ensemble with test-time augmentation on six radiographs outside the evaluation suite. \textbf{a,} Oblique finger. \textbf{b,} Lateral wrists. \textbf{c,} Heavily collimated, overexposed shoulder. \textbf{d,} Lateral ankle of an 11-year-old. \textbf{e,} Forearm with plate and screws. \textbf{f,} Lower leg of a 4-year-old. Panels a, c, and e are from the MURA validation split and are not among the MURA images annotated for training; b, d, and f are from Radiopaedia. Predictions use the released \method{} ensemble with test-time augmentation.}
\label{fig:appendix-failures}
\end{figure*}

\section{Failure cases}
\label{sec:appendix-failure-cases}
\Cref{fig:appendix-failures} illustrates limitations of \method{} on six radiographs outside the evaluation suite, including oblique and lateral views of peripheral anatomy, a heavily collimated and overexposed shoulder, pediatric anatomy, and surgical hardware. These examples highlight gaps in the acquisition conditions and anatomical variation represented by the training sources, motivating additional targeted training data.

\section{Code availability}
\label{sec:appendix-code-availability}

Code is available at \url{https://github.com/VictorButoi/FleXray}. Trained model weights are available at \url{https://huggingface.co/VictorButoi/flexray-base}. The release is intended for research use. Our enhanced-DRRs are distributed as a precomputed dataset in the same repository (\cref{sec:appendix-data-sources}).

\section{Data availability}
\label{sec:appendix-data-sources}

We used the following publicly available CT datasets (\cref{tab:methods-training-data}):
\begin{itemize}
    \item ElbowCT (\url{https://figshare.com/articles/dataset/3D_models_of_elbow_joints_along_with_corresponding_CT_data_from_Chinese_individuals/28245599})
    \item HaN-Seg (\url{https://han-seg2023.grand-challenge.org/})
    \item MOOSE ({\def\UrlBreaks{\do\/\do\-}\url{https://registry.opendata.aws/enhance-pet-1-6k/}})
    \item PedsCT (\url{https://www.cancerimagingarchive.net/collection/pediatric-ct-seg/})
    \item RSNAFrac (\url{https://www.kaggle.com/competitions/rsna-2022-cervical-spine-fracture-detection/})
    \item Shoulder-CT ({\def\UrlBreaks{\do\/\do\-}\url{https://www.kaggle.com/datasets/syxlicheng/automatically-transform-ct-datasets-into-drrs}})
\end{itemize}
We used the following publicly available X-ray datasets (\cref{tab:methods-training-data,tab:methods-eval-data}):
\begin{itemize}
    \item AASCE (\url{https://aasce19.github.io/})
    \item BTXRD (\url{https://doi.org/10.6084/m9.figshare.27865398})
    \item DarwinCVD19 ({\def\UrlBreaks{\do\/\do\-}\url{https://darwin.v7labs.com/v7-labs/covid-19-chest-x-ray-dataset}})
    \item DeepFluoro (\url{https://huggingface.co/datasets/eigenvivek/xvr-data})
    \item ElbowLat (\url{https://universe.roboflow.com/ionspace/elbow_lat-lnn0s-pmycd})
    \item FootBones (\url{https://universe.roboflow.com/monchbot1/foot_op})
    \item HandBones (\url{https://universe.roboflow.com/boneage-x90qt/-hand-bones-mdjkr})
    \item HipRay (\url{https://data.mendeley.com/datasets/zm6bxzhmfz/1})
    \item LowerLimbs (\url{https://universe.roboflow.com/orthopedicstitching/bone-identifier-1rey5})
    \item MTDDH (\url{https://doi.org/10.57760/sciencedb.24372})
    \item MURA (\url{https://stanfordmlgroup.github.io/competitions/mura/})
    \item RAM-W600 (\url{https://huggingface.co/datasets/TokyoTechMagicYang/RAM-W600})
    \item PedsTorso (\url{https://universe.roboflow.com/monchbot1/thoracoabdominal})
    \item VinDr-Rib (\url{https://vindr.ai/ribcxr})
\end{itemize}
Versions of the X-ray datasets whose licenses permit redistribution (BTXRD, ElbowLat, FootBones, HandBones, HipRay, LowerLimbs, and MTDDH; all CC BY 4.0), repackaged with dataset-native labels in the format that \method{} ingests directly, are available at \url{https://huggingface.co/datasets/VictorButoi/flexray-data}, together with the per-image split assignments and quality-control exclusions of every X-ray source. The same repository holds our manual forearm and humerus annotations for MURA (masks and an image manifest keyed to MURA file paths, without the source images, in accordance with the MURA Research Use Agreement). The remaining sources are not redistributed; for these we provide the label specifications used to map them into the protocol. Our 138,063 quality-controlled enhanced DRRs are synthetic and released with this work under CC BY-NC 4.0, inherited from the AutoPET-derived subset of the MOOSE source CTs~\cite{gatidis2022whole}, in the \texttt{FluXray/} folder of the same repository.

\section{Acknowledgments}
\label{sec:appendix-acknowledgments}

This work was supported by the National Science Foundation Graduate Research Fellowship Program, the MIT Health and Life Sciences Collaborative (HEALS), the MIT CSAIL METEOR Fellowship, an MGB Neuroscience Institute Transformative Scholar Award, Quanta Computer Inc., and NIH grants R01 EB033773, NIBIB 5T32EB001680-19, and S10 OD038222. Compute was also provided by the Massachusetts Life Sciences Center (MLSC).

\section{Author Contributions}
\label{sec:appendix-contributions}

V.I.B. developed the code, curated and preprocessed data, trained models, conducted experiments, created figures, and led manuscript development. V.G. developed the rendering toolkit central to the work, provided training and evaluation datasets, and executed and wrote the initial draft of the 2D-3D registration experiment. N.D. conceived the initial idea, identified the training and evaluation datasets, and provided technical direction throughout the project. All authors contributing to editing and providing technical insight and feedback on drafts.

\clearpage

\section{Appendix}
\label{sec:appendix}

\subsection{Statistics and reproducibility}
\label{sec:appendix-statistics}

\paragraph{Uncertainty.} Unless stated otherwise, results are means with 95\% percentile-bootstrap confidence intervals from 10,000 resamples. Per-dataset estimates resample images; equal-dataset summaries resample datasets and then images. Task-specific analyses resample inference seeds (TTA), cases (BTXRD), or matched label--frame observations (view-angle analyses). MTDDH repeats are averaged within images before resampling. Registration errors are summarized by medians.

\paragraph{Comparisons.} Paired \(t\)-tests assess mean differences in Dice and Cobb-angle error between methods evaluated on the same images; Cobb comparisons require measurable predictions from both methods. Welch's \(t\)-test compares attenuation-related improvements between independent healthy and pneumonia groups. Registration errors are compared with Wilcoxon signed-rank tests, using ranks to limit the influence of extreme error magnitudes. Success and failure rates are reported descriptively. All tests are two-sided at \(\alpha ={}\)0.05. Reported \textit{p} values are Holm-corrected across baseline comparisons or ablation pairs within datasets, within Cobb-angle severity strata, and across diagnostic conditions in the attenuation analysis.

\subsection{Implementation details}
\label{sec:appendix-implementation-details}

\method{} is trained using PyTorch~\cite{paszke2019pytorch} on NVIDIA H200 GPU nodes; each ensemble member trains on a single H200 for approximately 60--70 hours (roughly 320 GPU-hours for the ensemble). We use PyTorch automatic mixed precision (AMP) with bfloat16 autocasting for eligible forward-pass operations; model parameters and optimizer updates remain in FP32. We store our datasets in individual LMDB databases using \texttt{thunderpack}~\cite{ortiz2023thunderpack} in \texttt{numpy}~\cite{harris2020array} arrays. We store images and volumes as \texttt{float16} arrays and segmentation labels as \texttt{uint8} arrays. We generate projections from CTs using \texttt{nanoDRR}~\cite{gopalakrishnan2022fast} (\url{https://github.com/eigenvivek/nanodrr}). Standard data augmentations were implemented with the Kornia library~\cite{eriba2019kornia}. The MURA forearm and humerus subsets were annotated in CVAT~\cite{cvat2023} using SAM2~\cite{ravi2024sam} as an interactive point-prompted segmentation tool, followed by manual polygon correction of every mask.

\subsection{Anatomical coverage of public X-ray segmentation data}
\label{sec:appendix-coverage-provenance}

\Cref{fig:overview}b summarizes the anatomical coverage of the public X-ray segmentation datasets we located, listed in \cref{tab:appendix-coverage-datasets}. Each dataset's native labels are mapped into the 60-structure protocol as in \cref{sec:appendix-label-harmonization}, and an image counts toward a structure when it carries a mask of that structure, summed over the dataset's partitions. For display, the protocol is collapsed to 25 regions: the individual vertebrae form the spine, the individual ribs form the ribs, and the clavicles are omitted. The distal radius and ulna visible in hand and wrist radiographs are not counted as coverage of those bones. The skeleton is the surface of one MOOSE subject's label map, and each region is tinted by the logarithm of its image count on a shared color scale.

\begin{table}[h]
\centering
\scriptsize
\setlength{\tabcolsep}{4pt}
\renewcommand{\arraystretch}{1.15}
\caption{\textbf{Public X-ray segmentation datasets behind \cref{fig:overview}b.} Regions are the collapsed labels of \cref{sec:appendix-coverage-provenance} that each dataset annotates. Labeled X-rays is the number of images carrying a mask of at least one listed region, summed over the dataset's partitions. The radii and ulnae of the hand and wrist datasets are not counted because those radiographs show only the distal ends of the bones.}
\label{tab:appendix-coverage-datasets}
\begin{tabularx}{\columnwidth}{@{}l>{\raggedright\arraybackslash}Xr@{}}
\toprule
Dataset & Regions & Labeled X-rays \\
\midrule
CheXmask~\cite{gaggion2024chexmask} & Heart, lungs & 657,566 \\
\lightrule
DarwinCVD19~\cite{v7labs-covid19} & Lungs & 6,347 \\
\lightrule
MendeleyCXR~\cite{danilov2022chest} & Lungs & 704 \\
\lightrule
RAM-W600~\cite{yang2025ram} & Carpals, metacarpals & 618 \\
\lightrule
ElbowLat~\cite{ionspace-elbowlat} & Humeri, radii, ulnae & 601 \\
\lightrule
FootBones~\cite{FootBones} & Metatarsals, toes & 571 \\
\lightrule
DeepFluoro~\cite{grupp2020automatic} & Femurs, hips, sacrum, spine & 362 \\
\lightrule
VinDr-Rib~\cite{nguyen2021vindr} & Ribs & 245 \\
\lightrule
HipRay~\cite{gut2021x} & Femurs, hips & 139 \\
\lightrule
HandBones~\cite{HandBones} & Carpals, metacarpals, phalanges & 93 \\
\lightrule
PedsTorso~\cite{monchbot1_thoracoabdominal_2026} & Lungs, spine & 78 \\
\lightrule
LowerLimbs~\cite{bone-identifier-1rey5_dataset} & Femurs, fibulae, tibiae & 56 \\
\bottomrule
\end{tabularx}
\end{table}

\subsection{Radiopaedia image attributions}
\label{sec:appendix-radiopaedia-attributions}

The 18 radiographs shown in \cref{fig:robustness-eval}a and the three shown in \cref{fig:appendix-failures}b,d,f are drawn from Radiopaedia.org and are used under the Creative Commons BY-NC-SA 3.0 license, in accordance with the Radiopaedia image use and attribution guidelines (\url{https://radiopaedia.org/articles/using-and-attributing-images-from-radiopaedia-1}). Panels are numbered in reading order: band~1 is the upper pair of source and overlay rows and band~2 the lower pair, with columns counted left to right. Each source X-ray and its segmentation overlay share one attribution.

\noindent \radcite{\textbf{Panel 1} (band 1, column 1): Case courtesy of Bahman Rasuli, Radiopaedia.org, rID: 76252.}{Source: \href{https://prod-images-static.radiopaedia.org/images/52385720/a8ef731d274b273f1526d8d0ffbe3b.jpg}{Radiopaedia image 52385720}.}%
\radcite{\textbf{Panel 2} (band 1, column 2): Case courtesy of Andrew Murphy, Radiopaedia.org, rID: 48335.}{Source: \href{https://prod-images-static.radiopaedia.org/images/25366595/06213351c7e958afb1c15e81b56a8d.jpg}{Radiopaedia image 25366595}.}%
\radcite{\textbf{Panel 3} (band 1, column 3): Case courtesy of Frank Gaillard, Radiopaedia.org, rID: 37967.}{Source: \href{https://prod-images-static.radiopaedia.org/images/14074481/92b1c10a223ee8833c8a29924f68d9.jpg}{Radiopaedia image 14074481}.}%
\radcite{\textbf{Panel 4} (band 1, column 4): Case courtesy of Tan (Vivian) Hooi Hooi, Radiopaedia.org, rID: 200442.}{Source: \href{https://radiopaedia.org/cases/hepatopulmonary-fusion}{Radiopaedia case 200442}.}%
\radcite{\textbf{Panel 5} (band 1, column 5): Case courtesy of Tudor Hughes, Radiopaedia.org, rID: 214066.}{Source: \href{https://prod-images-static.radiopaedia.org/images/71451814/20a3b13237e04f1d25b975dce9cffbb5abcdfe6366bec746b9a3a08b410e4234.jpg}{Radiopaedia image 71451814}.}%
\radcite{\textbf{Panel 6} (band 1, column 6): Case courtesy of Sigmund Stuppner, Radiopaedia.org, rID: 45298.}{Source: \href{https://prod-images-static.radiopaedia.org/images/22567772/0929ec1ec7ff2002e7343047f46909.JPG}{Radiopaedia image 22567772}.}%
\radcite{\textbf{Panel 7} (band 1, column 7): Case courtesy of Liam Pugh, Radiopaedia.org, rID: 58094.}{Source: \href{https://prod-images-static.radiopaedia.org/images/35566320/be86e4ccd6e9c60882f3e6d023f384.jpg}{Radiopaedia image 35566320}.}%
\radcite{\textbf{Panel 8} (band 1, column 8): Case courtesy of Tudor Hughes, Radiopaedia.org, rID: 222397.}{Source: \href{https://prod-images-static.radiopaedia.org/images/72933366/5f52a272266d23df07001238501dae7a4b035c43a0f85ae7e89d618ba79ee3d0.jpg}{Radiopaedia image 72933366}.}%
\radcite{\textbf{Panel 9} (band 1, column 9): Case courtesy of Stefan Tigges, Radiopaedia.org, rID: 195402.}{Source: \href{https://prod-images-static.radiopaedia.org/images/67285607/58fdb647e4641434037d928a8b9056fd7bddb86146154d1c1666e9a3bbf39fef.jpeg}{Radiopaedia image 67285607}.}%
\radcite{\textbf{Panel 10} (band 2, column 1): Case courtesy of Andrew Dixon, Radiopaedia.org, rID: 31533.}{Source: \href{https://prod-images-static.radiopaedia.org/images/8689771/59e95d73ea42ba8a966115145b13ae.jpg}{Radiopaedia image 8689771}.}%
\radcite{\textbf{Panel 11} (band 2, column 2): Case courtesy of Yaïr Glick, Radiopaedia.org, rID: 89657.}{Source: \href{https://radiopaedia.org/cases/bone-within-a-bone-1}{Radiopaedia case 89657}.}%
\radcite{\textbf{Panel 12} (band 2, column 3): Case courtesy of Ian Bickle, Radiopaedia.org, rID: 46399.}{Source: \href{https://prod-images-static.radiopaedia.org/images/23887924/77d47c262581a4d93b835aac8f2856.jpg}{Radiopaedia image 23887924}.}%
\radcite{\textbf{Panel 13} (band 2, column 4): Case courtesy of Amanda Er, Radiopaedia.org, rID: 85561.}{Source: \href{https://prod-images-static.radiopaedia.org/images/54151097/bf08cac3156189e9301a1e31712eed2c2b65a20c132775bfef81079a07f2b107.png}{Radiopaedia image 54151097}.}%
\radcite{\textbf{Panel 14} (band 2, column 5): Case courtesy of Tudor Hughes, Radiopaedia.org, rID: 220698.}{Source: \href{https://prod-images-static.radiopaedia.org/images/72665895/4c2e169512efb6a663f7c8b8bc3def7f3201f2d44e31731443d4c26ebbc7cad9.jpg}{Radiopaedia image 72665895}.}%
\radcite{\textbf{Panel 15} (band 2, column 6): Case courtesy of Andrew Murphy, Radiopaedia.org, rID: 48335.}{Source: \href{https://prod-images-static.radiopaedia.org/images/25366594/489baf76895a943e59cc9f53336500.jpg}{Radiopaedia image 25366594}.}%
\radcite{\textbf{Panel 16} (band 2, column 7): Case courtesy of Stefan Tigges, Radiopaedia.org, rID: 195402.}{Source: \href{https://prod-images-static.radiopaedia.org/images/67285609/be7d16a9533d48303287a5c688f3c1ab35a23d12a05c6c4c8d12d9e1ac617a26.jpeg}{Radiopaedia image 67285609}.}%
\radcite{\textbf{Panel 17} (band 2, column 8): Case courtesy of Ian Bickle, Radiopaedia.org, rID: 46578.}{Source: \href{https://prod-images-static.radiopaedia.org/images/23961548/33913dda114390eb9917e9ac394873.jpg}{Radiopaedia image 23961548}.}%
\radcite{\textbf{Panel 18} (band 2, column 9): Case courtesy of Andrew Murphy, Radiopaedia.org, rID: 48227.}{Source: \href{https://prod-images-static.radiopaedia.org/images/25293680/89e4710c669675e3aa9a3b5da49f05.jpg}{Radiopaedia image 25293680}.}%
\radcite{\textbf{\Cref{fig:appendix-failures}b}: Case courtesy of Andrew Murphy, Radiopaedia.org, rID: 48227.}{Source: \href{https://prod-images-static.radiopaedia.org/images/25293683/81945cf07e8313a243b91f9ccc3f7a.jpg}{Radiopaedia image 25293683}.}%
\radcite{\textbf{\Cref{fig:appendix-failures}d}: Case courtesy of Tudor Hughes, Radiopaedia.org, rID: 220012.}{Source: \href{https://prod-images-static.radiopaedia.org/images/72575068/2928041364bd2167f82a9cd3477cbe3d725d1fae5ee31e216ff81e395cceb885.jpg}{Radiopaedia image 72575068}.}%
\radcite{\textbf{\Cref{fig:appendix-failures}f}: Case courtesy of Andrew Kirby, Radiopaedia.org, rID: 238301.}{Source: \href{https://prod-images-static.radiopaedia.org/images/75212780/dr-original.jpg}{Radiopaedia image 75212780}.}%

\end{document}